\documentclass[acmtog,screen,authorversion]{acmart}

\usepackage{booktabs} 

\usepackage[ruled]{algorithm2e} 
\usepackage{makecell}
\usepackage{amsmath}

\SetAlFnt{\small}
\SetAlCapFnt{\small}
\SetAlCapNameFnt{\small}
\SetAlCapHSkip{0pt}

\setcopyright{acmcopyright}
\setcopyright{acmlicensed}
\setcopyright{rightsretained}
\setcopyright{usgov}
\setcopyright{usgovmixed}
\setcopyright{cagov}
\setcopyright{cagovmixed}

\newcommand{\head}[1]{{\vspace{2.5mm}\noindent\textbf{#1}}}
\usepackage{enumitem}

\begin{document}
\title{PS-CAD: Local Geometry Guidance via Prompting and Selection for CAD Reconstruction}

\author{Bingchen Yang}
\author{Haiyong Jiang}
\author{Hao Pan}
\author{Peter Wonka}
\author{Jun Xiao}
\author{Guosheng Lin} 



\begin{abstract}
Reverse engineering CAD models from raw geometry is a classic but challenging research problem. 
In particular, reconstructing the CAD modeling sequence from point clouds provides great interpretability and convenience for editing.
Analyzing previous work, we observed that a CAD modeling sequence represented by tokens and processed by a generative model, does not have an immediate geometric interpretation.
To improve upon this problem, we introduce geometric guidance into the reconstruction network. 
Our proposed model, PS-CAD, reconstructs the CAD modeling sequence one step at a time as illustrated in Fig.~\ref{fig:teaser}. At each step, we provide two forms of geometric guidance. First, we provide the geometry of surfaces where the current reconstruction differs from the complete model as a point cloud. This helps the framework to focus on regions that still need work. Second, we use geometric analysis to extract a set of planar prompts, that correspond to candidate surfaces where a CAD extrusion step could be started. Our framework has three major components. Geometric guidance computation extracts the two types of geometric guidance. Single-step reconstruction computes a single candidate CAD modeling step for each provided prompt. Single-step selection selects among the candidate CAD modeling steps. The process continues until the reconstruction is completed.
Our quantitative results show a significant improvement across all metrics. For example, on the dataset DeepCAD, PS-CAD improves upon the best published SOTA method by reducing the geometry errors (CD and HD) by {$10\%$}, and the structural error (ECD metric) by about {$15\%$}.
\end{abstract}

\begin{teaserfigure}
  \centering
  \includegraphics[width=\textwidth]{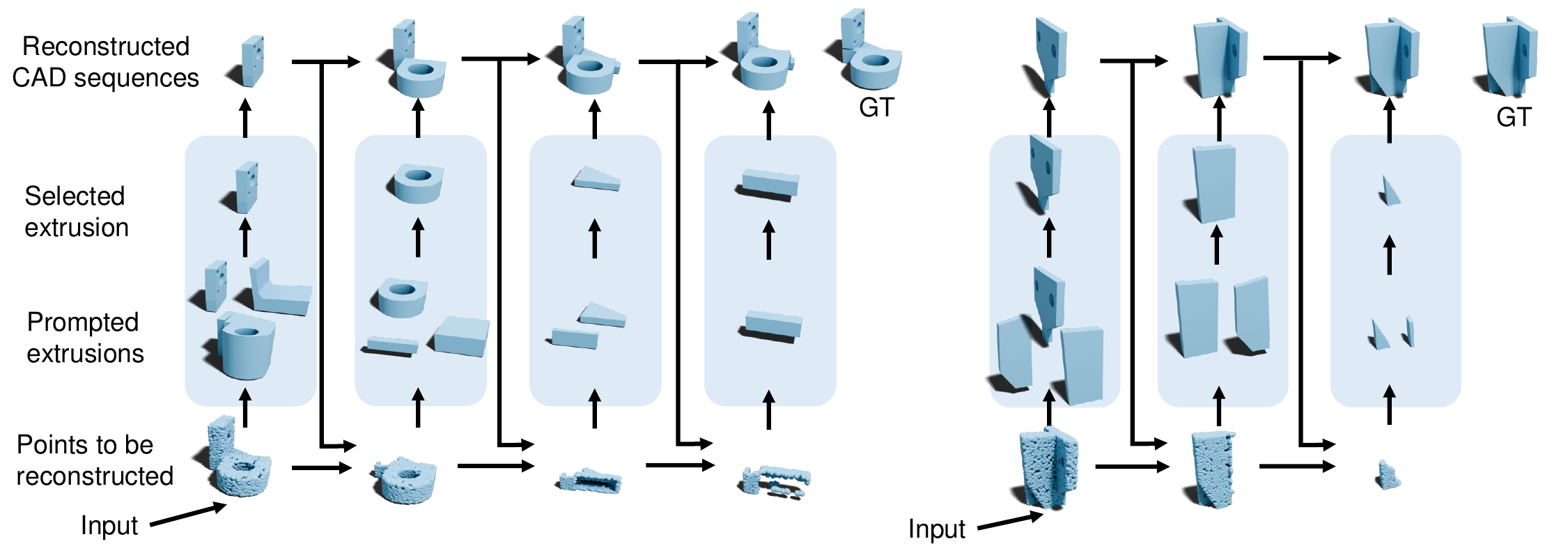}
  \caption{ Two examples for the iterative PS-CAD reconstruction.
  PS-CAD takes a point cloud as input (bottom left labeled input) and produces a CAD modeling sequence (the top row) using an iterative reconstruction architecture. 
  For each iteration (one column in the figure), the method first determines a point cloud highlighting the parts that still need the most work (bottom row labeled points to be reconstructed). Then we extract planar prompts to guide the estimation of extrusion cylinders (middle labeled prompted extrustions). After that, the method selects the most suitable one (middle labeled selected extrusion). The top shows the reconstructed CAD sequence step by step. This process iterates until the final shape is reconstructed.
  }
  \Description{}
  \label{fig:teaser}
\end{teaserfigure}

%
%


%
%


\maketitle

\section{Introduction}
\label{sec:intro}
    Reconstructing Computer-Aided Design (CAD) models from scans is one of the long-sought goals in geometric modeling and plays a critical role in reverse engineering.
    In addition to 3D surface details, CAD models also capture the structure of the 3D models which facilitates 3D editing.
        
\begin{figure*}[t]
    \centering
    \includegraphics[width=\textwidth]{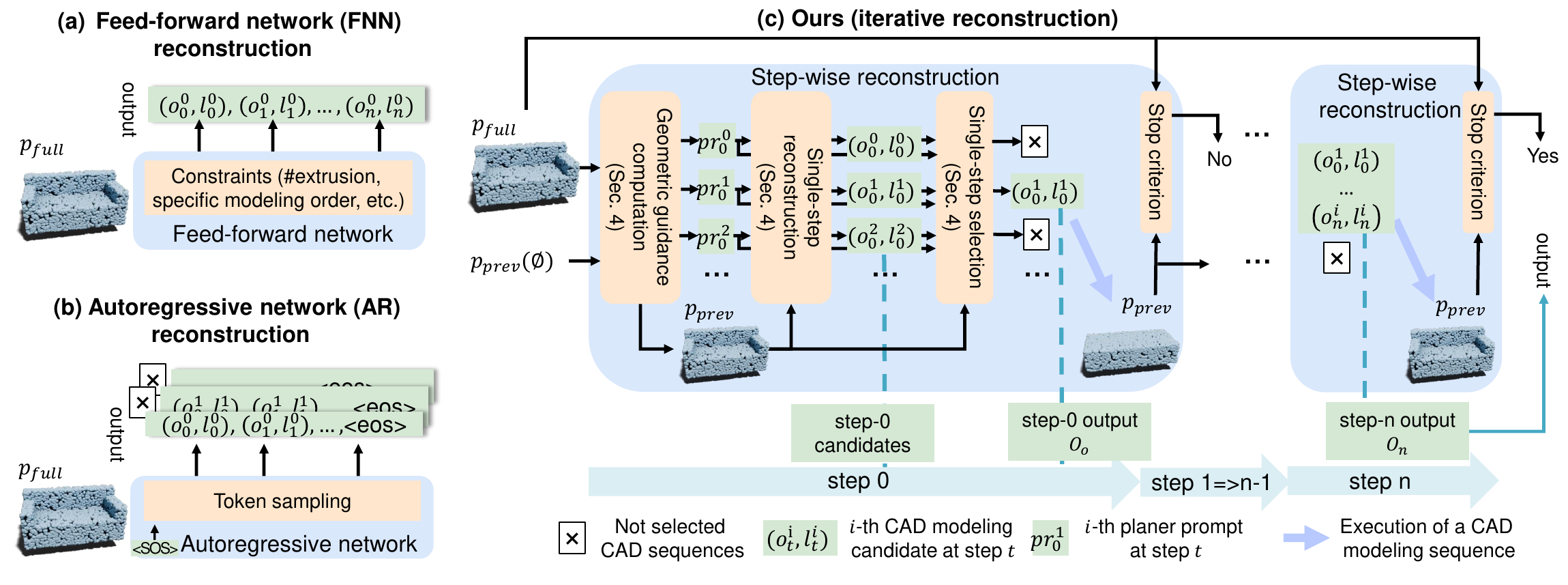}
    \caption{Comparison of different reconstruction pipelines: (a) feed-forward network (FFN), (b) autoregressive network (AR), and (c) our iterative reconstruction. In (a), the FFN decomposes an input point cloud into a fixed number of extrusion cylinders and their Boolean operations, each of which is represented with a CAD modeling step $(o_t^0, l_t^0)$. 
    In (b), the AR reconstructs the CAD modeling sequence token-by-token. By sampling different tokens without specific geometric meaning, the AR can reconstruct multiple CAD modeling sequences and output the one with the lowest reconstruction error. In (c), we propose an iterative pipeline with each iterative step consisting of three components: geometric guidance computation,  single-step reconstruction, and single-step selection. At step $t$, the geometric guidance computation generates candidate prompts $pr_t^i$ (in the form of sampled planes) and local geometric information $p_\text{ref}$ to provide geometric guidance to the single-step reconstruction module. The single-step reconstruction module produces one candidate CAD modeling step $(o_t^i, l_t^i)$. 
    Then, the single-step sequence selection module selects the CAD modeling step with the highest scores and concatenates the corresponding tokens to the output sequence. This process iterates $n$ steps until a stop criterion is satisfied.}
    \label{fig:pip-overview}
\end{figure*}

    Many endeavors have been devoted to this problem. 
    Some approximate an object with a set of parametric primitives with a focus on shape abstraction~\cite{tulsiani2017learning, DBLP:conf/cvpr/LiSDYG19, DBLP:conf/iccv/YanYMHVH21,DBLP:conf/cvpr/PaschalidouUG19} or detail preservation~\cite{DBLP:conf/nips/0005X0TZM020,DBLP:journals/tog/GuoLPLTG22,DBLP:conf/siggraph/LiLYGG023,DBLP:conf/cvpr/Zhu00ZN023}. However, simply assembling parametric primitives provides only the surface and boundary information of the shape. These approaches do not capture the structure of the CAD model encoded by modeling operations.
    To this end, Constructive Solid Geometry (CSG)-based representations~\cite{DBLP:conf/cvpr/SharmaGLKM18, DBLP:journals/pami/SharmaGLKM22, DBLP:conf/nips/EllisNPSTS19, DBLP:conf/cvpr/JonesWR22, DBLP:conf/nips/KaniaZK20, DBLP:conf/iccv/RenZ0LJCZPZZY21, DBLP:conf/eccv/RenZCLZ22, DBLP:conf/cvpr/YuCLSSMZ22, DBLP:conf/nips/YuCTMZ23} are explored for their explicit modeling steps. 
\begin{figure}
    \centering
    \vspace{-1mm}
    \includegraphics[width=0.9\linewidth]{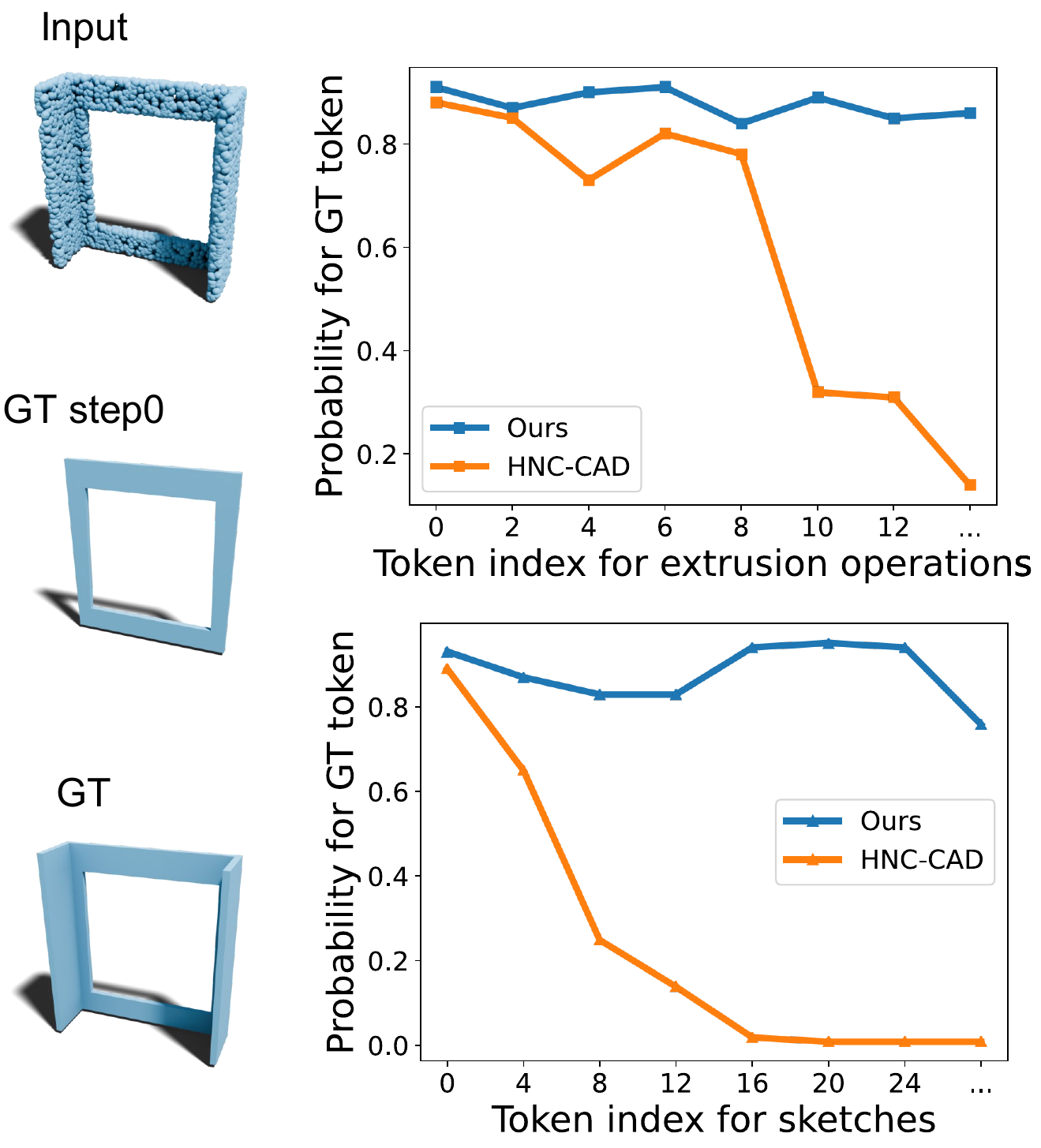}
    \caption{Token probabilities for our method and HNC-CAD~\cite{DBLP:conf/icml/XuJLWF23}. Left: the input point cloud, the first CAD modeling step of the ground-truth, and the final output of the ground truth. Right: Probabilities of the ground truth tokens for sketches and extrusion operations for HNC-CAD and our method. Our method performs stably for each token while for HNC-CAD the probabilities drastically decrease after generating several tokens.}
    \label{fig:intro}
\end{figure}
    Another line of works
    ~\cite{DBLP:conf/iccv/WuXZ21,DBLP:conf/cvpr/UyCSGLBG22, ma2023multicad, DBLP:conf/cvpr/LiG0Y23, DBLP:conf/siggrapha/LambourneWJZSM22, khan2024cadsignet} take a step further and directly learn to estimate a CAD modeling sequences from an input CAD model.
    These works decompose a CAD model into a list of CAD modeling steps. In our work, we adopt a domain-specific language (DSL) for CAD~\cite{xu2022skexgen}, where a CAD modeling step consists of a sketch (a 2D curve), an extrusion, and a Boolean operation (union or subtraction). The 3D shape defined by a 2D sketch and extrusion is called an extrusion cylinder. Even though previous work may have some slight deviations from this model, we use it as a common framework to structure the discussion of existing literature. 
    There are two types of architectures used in previous work to tackle this problem. The first architecture is to generate CAD models using a feed-forward network (See Fig.~\ref{fig:pip-overview}-a, e.g.~\cite{ DBLP:conf/iccv/WuXZ21,DBLP:conf/cvpr/UyCSGLBG22,  ma2023multicad, DBLP:conf/cvpr/LiG0Y23, DBLP:conf/siggrapha/LambourneWJZSM22}). 
    The feed-forward network predicts a fixed number of extrusion cylinders and Boolean operations jointly all at once, thus lacking the flexibility to adapt to input geometry with a varying number of CAD modeling steps and to sample different CAD modeling designs. 
    The second architecture is to sequentially generate a CAD sequence using an auto-regressive deep learning model. (See Fig.~\ref{fig:pip-overview}-b, e.g.~\cite{xu2022skexgen, DBLP:conf/icml/XuJLWF23, khan2024cadsignet}). This model generates a CAD model one token at a time. A sequence of tokens (about 20 - 80) yields a single CAD modeling step.
    As each token does not have a special geometric meaning, this scheme does not consider local geometric contexts and results in a rapid descent in the softmax probability of the sampled tokens, indicating inaccurate token sampling during CAD reconstruction (See Fig.~\ref{fig:intro}).
    We further evolve this approach to yield a third type of architecture, generating a sequence one CAD modeling step at one time (See Fig.~\ref{fig:pip-overview}-c). We call this a prompt-and-select architecture. Each iteration of our framework outputs a single CAD modeling step and consists of three components: a) geometric guidance computation, b) single-step reconstruction, and c) single-step selection. We visualize the two previous network architectures and our proposed framework in Fig.~\ref{fig:pip-overview}. Step-wise CAD reconstruction is shown in Fig.~\ref{fig:teaser}.

    This enables us to improve upon existing auto-regressive models by providing local geometric guidance. An auto-regressive model obtains the current state as a sequence of already-generated tokens. This sequence of tokens has no immediate geometric interpretation and this requires a lot of processing to obtain a geometric interpretation before the model can reason about which token to output next. We propose to provide three types of local geometric guidance for the auto-regressive model to improve its functionality.
    First, we provide the difference between the current reconstruction and the input model as point cloud. This let's the model know which parts it should focus on. 
    Second, we geometrically analyze the model to find candidate planes from which a new CAD modeling step can be started. We then provide these candidate planes as input in the form of prompts, analog to segmentation prompts popular in recent work~\cite{DBLP:conf/iccv/KirillovMRMRGXW23}. This provides the context for generating the next candidate CAD modeling step and the generated sketches (curves) are encouraged to lie on the prompt planes.
    Third, we implement a geometric form of lookahead sampling. Instead of simply sampling one token at a time based on the computed probability, we generate multiple complete candidate CAD modeling steps (each encoded as a sequence of tokens) and then evaluate the candidates to choose the most suitable one. The network trained for this component is supervised with geometric criteria. Fig.~\ref{fig:teaser} shows how the proposed modeling pipeline builds a CAD model step by step. Fig.~\ref{fig:pip-overview}-c shows the proposed network architecture.

     Our main contributions are:    
    \begin{itemize}[noitemsep, leftmargin=5mm]
        \item We propose an iterative prompt-and-select architecture to progressively reconstruct the CAD modeling sequence of a target shape.
        \item We propose the concept of local geometric guidance for auto-regressive models and propose three ways to integrate this guidance into the auto-regressive framework.
        \item  Quantitative experiments demonstrate a significant improvement over the current state of the art. For example, on the dataset DeepCAD, our framework PS-CAD improves upon the best published SOTA method by reducing the geometry errors (CD and HD) by $10\%$, and the structural error (ECD) by about $15\%$.
    \end{itemize}

\section{Related Work}
\label{sec:related works}
    In this section, we review related works on 3D CAD reconstruction with different output formats, ranging from parametric primitives, and CSG trees, to the CAD modeling sequence.

    \head{CAD Primitive Reconstruction.}
    Most approaches in the scheme represent an input 3D shape as a set of parametric primitives to be assembled, where the parametric primitives can be edges, surface patches, 3D solids, etc.
    Traditional approaches first segment and fit an input shape with basic parametric primitives (e.g. planes, spheres, and cylinders)~\cite{DBLP:journals/cgf/SchnabelWK07, DBLP:journals/tog/KimLMCDF13, DBLP:journals/tog/Sorkine-HornungM14}. 
    Recent advances in learning-based methods have improved the robustness of primitive fitting approaches~\cite{tulsiani2017learning, DBLP:conf/cvpr/LiSDYG19, DBLP:conf/iccv/YanYMHVH21} by being tolerable to outliers, meanwhile allowing for a wider variety of primitives for enhanced shape representation and better performance, including e.g. super quadratics~\cite{DBLP:conf/cvpr/PaschalidouUG19}, convexes~\cite{DBLP:conf/cvpr/DengGYBHT20}, learnable B\'{e}zier~\cite{DBLP:conf/ijcai/FuWLXA23} and B-spline patches~\cite{DBLP:conf/eccv/SharmaLMKCM20}, implicit fields~\cite{DBLP:conf/cvpr/ParkFSNL19}, etc. 
    Another category of approaches~\cite{DBLP:conf/nips/0005X0TZM020,DBLP:journals/tog/GuoLPLTG22,DBLP:conf/siggraph/LiLYGG023,DBLP:conf/cvpr/Zhu00ZN023} emphasize the preservation of sharp geometric features and learn to fit edges, corners that have a parity relation with surface patches from the input.
    Though these works can produce high-quality geometry, their results are mainly primitives missing the high-level CAD modeling operations, and therefore not easy to apply further editing.  
    
    \head{{CSG-tree based CAD Reconstruction.}}
    Pioneering works~\cite{DBLP:journals/cad/ShapiroV91, DBLP:journals/tog/ShapiroV93, DBLP:journals/cad/BucheleC04, DBLP:conf/gecco/HamzaS04, DBLP:journals/cad/FayolleP16} investigate to express input 3D shape as Constructive Solid Geometry (CSG)~\cite{DBLP:conf/siggraph/LaidlawTH86} with Boolean operations, well maintaining the shape construction information without loss of expressiveness. 
    Typically, search-based algorithms and pruning strategies are applied to build CSG trees \cite{DBLP:journals/tog/DuIPSSRSM18, DBLP:journals/cgf/WuXW18}. 
    For learning-based CSG reconstruction, the key challenge is to combine the indifferentiable CSG tree construction with differentiable primitive fitting.
    Previous works address this problem by using reinforcement learning~\cite{DBLP:conf/cvpr/SharmaGLKM18, DBLP:journals/pami/SharmaGLKM22,DBLP:conf/nips/EllisNPSTS19,DBLP:conf/cvpr/JonesWR22}, recursive constructions based on hand-crafted geometric rules~\cite{DBLP:journals/tog/GuoLPG22}, and novel learnable Boolean layers~\cite{DBLP:conf/nips/KaniaZK20}. 
    Successive approaches specify the CSG trees with fixed order and depth~\cite{DBLP:conf/iccv/RenZ0LJCZPZZY21,DBLP:conf/eccv/RenZCLZ22}, together with dedicated differentiable objectives and networks~\cite{DBLP:conf/cvpr/YuCLSSMZ22, DBLP:conf/nips/YuCTMZ23}. 
    However, to replicate the overall structure and retain fine geometric details with a fixed set of primitives, the structures of composed intermediate parts are seldom considered and therefore less interpretable and editable. 

    \head{{CAD Modeling sequence Reconstruction.}}
    To accommodate the modern CAD design pipeline, several approaches predict a CAD modeling sequence from a 3D CAD model. 
    DeepCAD~\cite{DBLP:conf/iccv/WuXZ21} encodes CAD modeling sequences as CAD domain-specific language (DSL) sequences and trains a transformer model conditioned on input to generate or reconstruct CAD DSL sequences directly.
    ZoneGraph~\cite{DBLP:conf/cvpr/XuPCWR21} employs a zone graph to represent the CAD model given as B-rep. A zone corresponds to a 3D entity derived from the B-rep faces. 
    In this way, predicting the CAD modeling sequence turns into a combinatorial problem in the discrete space permitted by the zone graph. 
    Leveraging well-annotated datasets~\cite{DBLP:journals/tog/WillisPLCDLSM21, DBLP:conf/iccv/WuXZ21}, Point2Cyl~\cite{DBLP:conf/cvpr/UyCSGLBG22} derives a set of parametric extrusion cylinders with a specified set size, based on segmentation of input shape.
    Lambourne et al.~\shortcite{DBLP:conf/siggrapha/LambourneWJZSM22} train a network to predict sketches and extruded values with a predefined CAD modeling sequence template. SECAD-Net~\cite{DBLP:conf/cvpr/LiG0Y23} reconstructs parametric extrusions and combines them as the input CAD model using Union operation. The closest work to ours is the concurrent CAD-SIGNet~\cite{khan2024cadsignet}. It reconstructs the CAD modeling sequence using an autoregressive network. As the core design, it uses predicted extrusion operation to assist in predicting the sketch.
    Our approach has the following differences from CAD-SIGNet: (1) We iteratively reconstruct the CAD modeling steps that splice to the complete CAD modeling sequence, rather than output the complete CAD modeling sequence (see Fig.~\ref{fig:pip-overview} (b) and (c) for comparison); (2) In each step we reconstruct multiple candidate CAD modeling steps based on prompts rather than token sampling (3) We devise a single-step selection module to select the best-fitted candidate, rather than rely on geometric criteria for CAD modeling sequence selection. 

\section{Overview} \label{ov}
\begin{figure}
    \centering
    \includegraphics[width=0.9\linewidth]{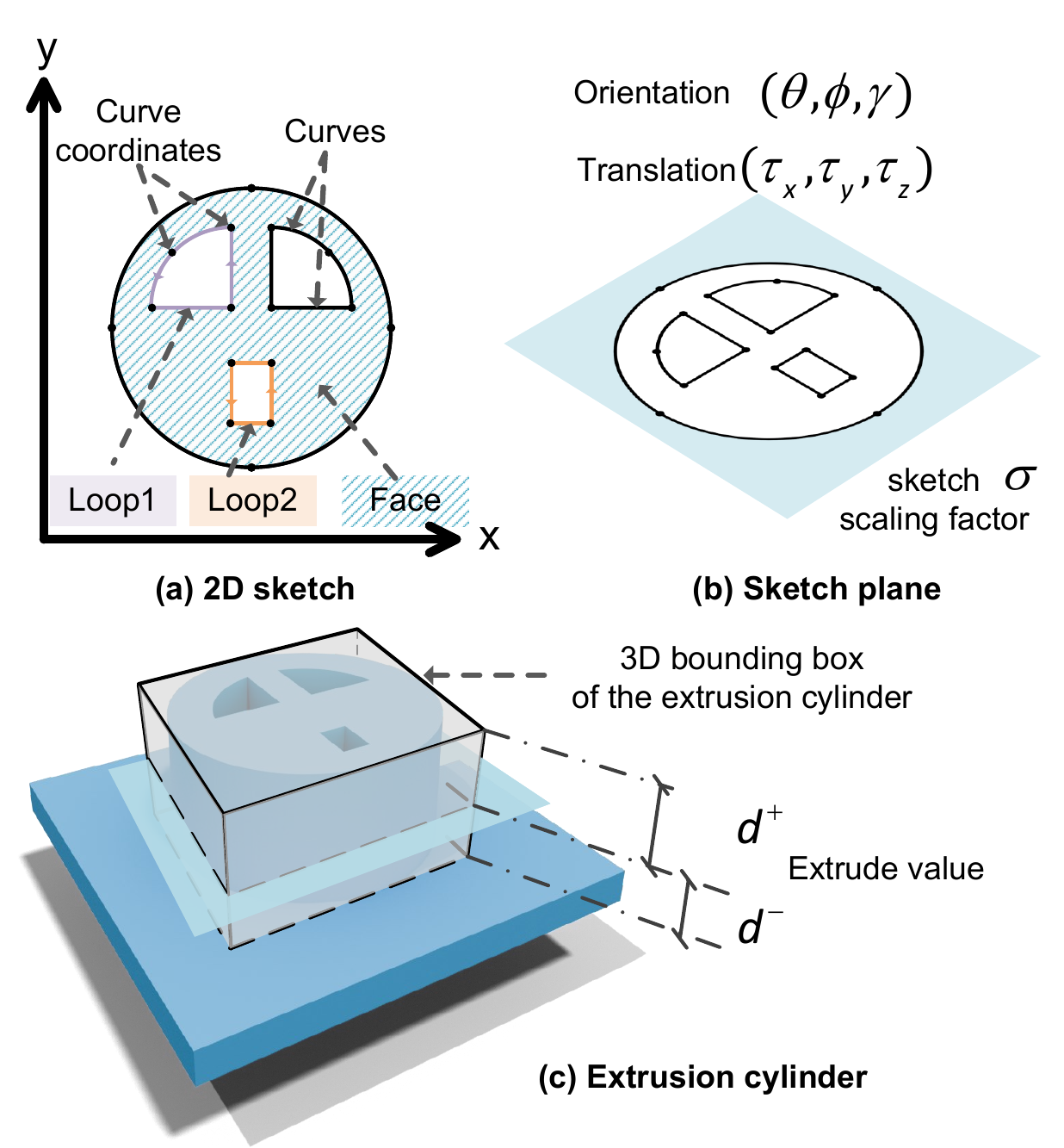}
    \caption{Illustration for the sketch and extrude representation. a) We show a 2D sketch consisting of four loops (three inner loops and one outer loop). The four loops define a face shaded in blue. Each of the loops consists of one or multiple curves, where each curve is a linear segment (defined by two points), an arc (defined by three points), or a circle (defined by four points). The outer loop is an example of a circle. This example shows a single face, but in general, a sketch with multiple faces is also allowed. b) The 2D sketch is transformed from its local 2D coordinate system to the 3D coordinate system of the CAD model by translation, orientation, and scaling. c) The 2D sketch is extruded from a height $d^-$ to a height $d^+$ defining an extrusion cylinder. We also use the bounding box of the extrusion cylinder in our computation.}
    \label{fig:DSL}
\end{figure}
    In this section, we first introduce the domain-specific language (DSL) used for encoding CAD modeling sequences in this paper. Then we give an overview of our approach. Our approach takes a 3D CAD model given as point cloud as input and outputs a DSL sequence.

\subsection{Preliminaries on the CAD DSL}\label{subsec:CAD-DSL}
    We use the CAD DSL from SkexGen~\cite{xu2022skexgen}. It decomposes a 3D CAD model into a sequence of CAD modeling steps $(o_t, l_t)$, where $o_t$ is an \emph{extrusion cylinder} and $l_t$ is a \emph{Boolean operation} (union or subtraction). An extrusion cylinder consists of a 2D sketch $s_t$ and an extrusion operation $e_t$.
    The CAD modeling sequence $O$ is a sequence of CAD modeling steps:
    \begin{equation}
        O =  [ (o_t, l_t)  ], \text{where } o_t = (e_t, s_t); l_t \in \{0, 1\}; t=0,1,\ldots,N_O-1, 
    \label{eq:CADSeq}
    \end{equation}
    where $N_O$ is the number of CAD modeling steps. We use $O_T$ to define the construction of the CAD model up until step $T$, i.e., the current state achieved by executing all $(o_t, l_t)$ with $t \leq T$.
    Next, we explain how $o_t$ is described by a 2D sketch $s_t$ and extrusion $e_t$.
    The sketch $s_t$ is organized as a "sketch-face-loop-curve" hierarchy. A sketch can contain one or more faces, where a face is a 2D region bounded by one or multiple loops. A loop is a closed 2D path consisting of one or more curves. A curve is either a line segment, an arc, or a circle. See Fig.~\ref{fig:DSL} for an example.
    A sketch $s_t$ is represented as a nested sequence:
    \begin{equation}
        s_t = \Big[(c_t^0, \varepsilon_c), \varepsilon_l, \ldots, \varepsilon_l, \ldots, \varepsilon_l, \ldots, \varepsilon_l, \varepsilon_f, \varepsilon_s\Big],
    \label{eq:CADSeq-s}
    \end{equation} 
    where $\varepsilon_c$, $\varepsilon_l$, $\varepsilon_f$, $\varepsilon_s$ denote the tokens to mark the end of a curve, loop, face, and sketch, respectively.
    $c_t^0$ is a list of four 2D coordinates $[(\zeta_x, \zeta_y), ...]\in\mathbb{R}^2$ to represent the outer circle in Fig.~\ref{fig:DSL}. To distinguish between the three types of curves, a line segment is described by 2 points, an arc by 3, and a circle by 4.
    The extrusion operation $e_t$ transforms the sketch $s_t$ into an extrusion cylinder $o_t$:
    \begin{equation}
        e_t = [d^+;d^-;\tau_x;\tau_y;\tau_z;\theta;\phi;\rho;\sigma;\varepsilon_e].
    \label{eq:CADSeq-e}
    \end{equation}
    Specifically, the 2D sketch is scaled with a factor of $\sigma\in\mathbb{R}$ and then transformed to the sketch plane with orientation $[\theta, \phi, \psi]\in\mathbb{R}^3$ and translation $[\tau_x, \tau_y, \tau_z]\in\mathbb{R}^3$. 
    $[d^-, d^+]$ represents the extruded distances along (or opposite) the sketch plane normal. 
    An illustration is shown in Fig.~\ref{fig:DSL}.
    Each extrusion cylinder $o_t$ is combined with the previous state $O_{t-1}$ using Boolean operation $l_t$, where $l_t$ is $1$ for the Union or $0$ for Subtraction.
    In our implementation, we adopt the discrete tokenization in SkexGen~\cite{xu2022skexgen} as the CAD DSLs to facilitate further autoregressive model learning. 
    Additional details on the DSL tokenization and CAD modeling sequences can be found in Supp.~\ref{app:temb}.

\subsection{Pipeline Overview}\label{subsec:pipeline}
\begin{figure}[t]
    \centering
    \vspace{-1mm}
    \includegraphics[width=\linewidth]{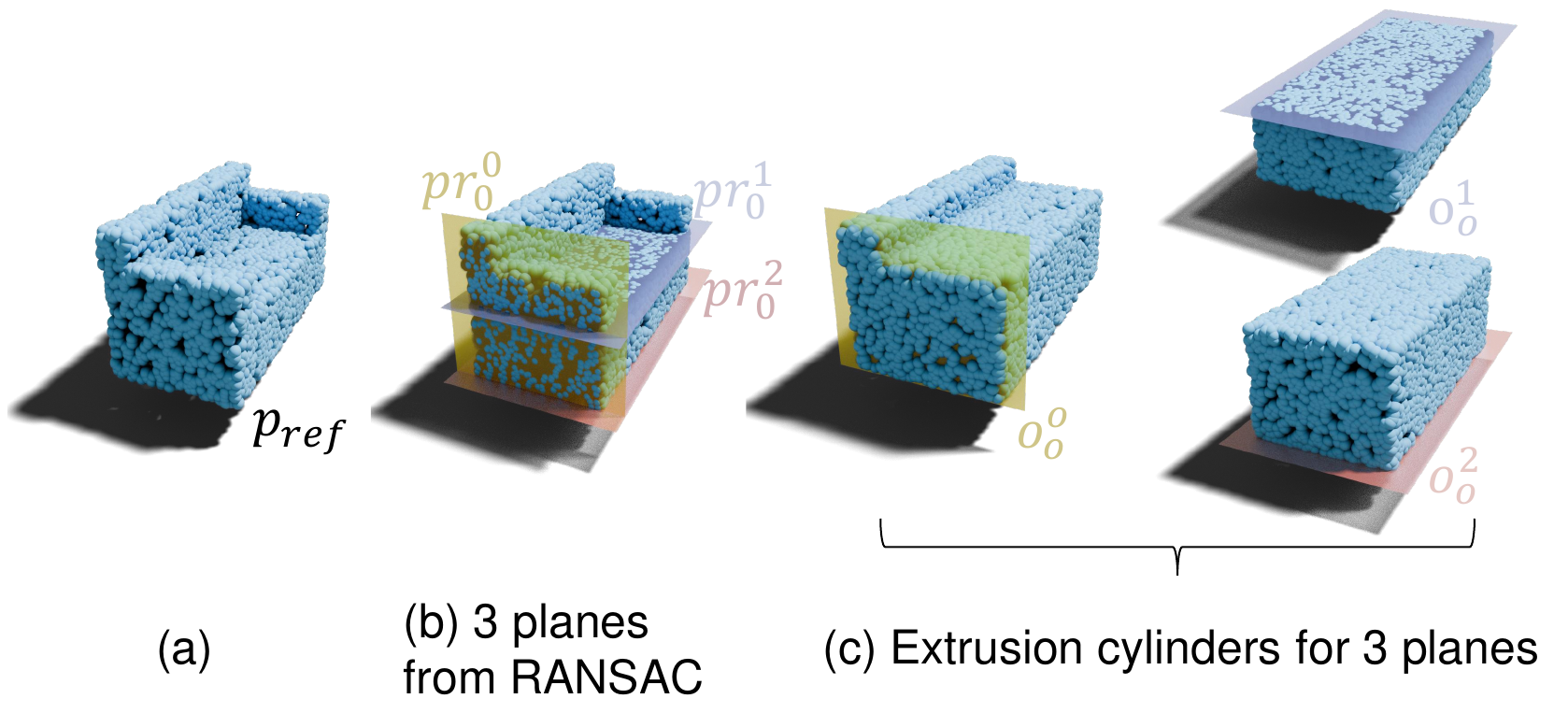}
    \caption{An illustration of $p_\text{ref}$ for the first step in Fig.~\ref{fig:pip-overview}, 3 detected planes from RANSAC, and their corresponding extrusion cylinders $o_0^* = (s^*_0, e_0^*)$. Only 3 detected planes are shown for simplicity.}
    \label{fig:extrusion4planes}
\end{figure}
    Given an input 3D CAD model as point cloud $p_\text{full} \in \mathbb{R}^{N_p \times 3}$ consisting of $N_p$ points, we learn to reconstruct an output CAD modeling sequence $O$ encoded by the CAD DSL.
    To tackle the problem, we present an iterative prompt-and-select pipeline as illustrated in Fig.~\ref{fig:pip-overview}.
    For each iteration $t$, the method executes the following algorithmic steps. 

    {\bf Geometric Guidance Computation:}
    We can analyze the previous state defined by CAD modeling sequence $O_{t-1}$ and point cloud $p_\text{full}$ to compute two forms of geometric guidance. First, we make use of the fact that extrusion cylinders have a planar bottom and top surface. We detect a set of candidate planes that are possible starting points for the next CAD modeling step $(o_t, l_t)$. These planes are encoded as planar prompts $pr_t^i$. Further, we compute a point cloud $p_\text{ref}$ that encodes the difference between $p_\text{full}$ and the previous state $O_{t-1}$ as additional geometric guidance. We encode $O_{t-1}$ by sampling it as point cloud $p_\text{prev} \in \mathbb{R}^{N_p \times 3}$.

    {\bf Single-Step Reconstruction:} 
    The single-step reconstruction module at step $t$ takes as inputs the planar prompts $pr_t^i$, the target point cloud $p_\text{full}$ and the point cloud describing the previous state $p_\text{prev}$.
    We compute a possible CAD modeling step $(o_t^i,l_t^i)$ for each sampled prompt $pr_t^i$ (see the illustration in Fig.~\ref{fig:extrusion4planes}). 
    %
    
    {\bf Single-Step Selection:}
    The single-step selection module at step $t$ takes all candidate CAD modeling steps $(o_t^i,l_t^i)$ as input. We evaluate the fitness of each CAD modeling step and select the one with the highest fitness. For example, the CAD modeling step $(o_0^1,l_0^1)$ is chosen in the step-0 of Fig.~\ref{fig:pip-overview}(c). Then, we concatenate the chosen CAD modeling step $(o_t,l_t)$ to the previous state $O_{t-1}$ and continue with the next iteration.

    The method iteratively executes the above three modules until a stop criterion is achieved, e.g., the reconstruction error is small enough. The final output is denoted as $O$.

\subsection{Common Modules}
    \head{Point cloud encoder.}
    We require a building block that can take an arbitrary point cloud and convert it into a set of tokens (features). We use the popular framework Point-MAE~\cite{DBLP:conf/eccv/PangWTLTY22} to obtain point cloud features.
    We pre-train Point-MAE based on a point cloud reconstruction task following the official repo\footnote{https://github.com/Pang-Yatian/Point-MAE} with input point cloud $p_\text{full}$ on the training dataset. 
    Note that this pre-training is independent of the DSL sequence reconstruction. Once Point-MAE is trained, we only keep the Point-MAE encoder for point cloud features extraction.
    During training, we froze the first three layers and finetuned the last three layers so that the network could be adapted for our specific tasks. 
    The point cloud encoder can encode an input point cloud $p$ to a set of tokens (features) $f$. For example, in later sections, we use $f_\text{full}, f_\text{prev}$, and $f_\text{ref}$ to denote the features computed for $p_\text{full}, p_\text{prev}$, and $p_\text{ref}$, respectively.

\section{Geometric Guidance Computation}\label{sec:cpg}
    The geometric guidance computation module at iteration (time step) $t$ takes the point cloud of the complete input CAD model $p_\text{full}$ as input. As additional input, we execute the token sequence describing the previous state $O_{t-1}$ and sample the resulting CAD model with point cloud $p_\text{prev}$. As output we compute a set of candidate prompts $pr_t^i$.

    \head{Point Cloud Segmentation}
    First, we would like to identify the distinct regions between $p_\text{full}$ and $p_\text{prev}$.
    The distinct region of $p_\text{full}$ compared to $p_\text{prev}$ represents the geometry not yet covered by $p_\text{prev}$, while the distinct region of $p_\text{prev}$ compared to $p_\text{full}$ contains auxiliary points in the construction that will no longer be present in the final model.
    For example, given an observation in Fig.~\ref{fig:ssr}(b), $p_\text{full}$ represents a sofa model, and $p_\text{prev}$ represents its base. The distinct region of $p_\text{full}$ corresponds to the armrest of the sofa. 
    However, point clouds for the bottom face of the armrest are absent in $p_\text{full}$ since it is sampled from the surface of the sofa, and the missing point cloud is complemented by the distinct region of $p_\text{prev}$.
    We train a segmentation network to predict binary masks $\mathcal{M}$ on $p_\text{full}$ and $p_\text{prev}$, labeling points with $1$ if they belong to a distinct region.
    The segmentation mask $\mathcal{M}$ is then applied to $p_\text{full}$ and $p_\text{prev}$ to get distinct point cloud regions $p_\text{ref}$. We use the notation $\mathcal{M}_\text{full}( p_\text{full})$ and $\mathcal{M}_\text{prev}( p_\text{prev})$ to denote an operation that returns the points with label $1$ from the point cloud $p_\text{full}$ and $p_\text{prev}$, respectively. The point cloud $p_\text{ref}$ is simply the concatenation of $\mathcal{M}_\text{full}( p_\text{full})$ and $\mathcal{M}_\text{prev}( p_\text{prev})$.

    \head{Plane Detection} We detect a set of candidate planes with an off-the-shelf RANSAC method and encode them as planar prompt $pr_t^i$. A planar prompt is encoded by randomly sampling 64 inlier points from a plane detected by RANSAC.

\section{Single-step Reconstruction}\label{sec:ssr}
\begin{figure*}
    \centering
    \vspace{-3mm}
    \includegraphics[width=\textwidth]{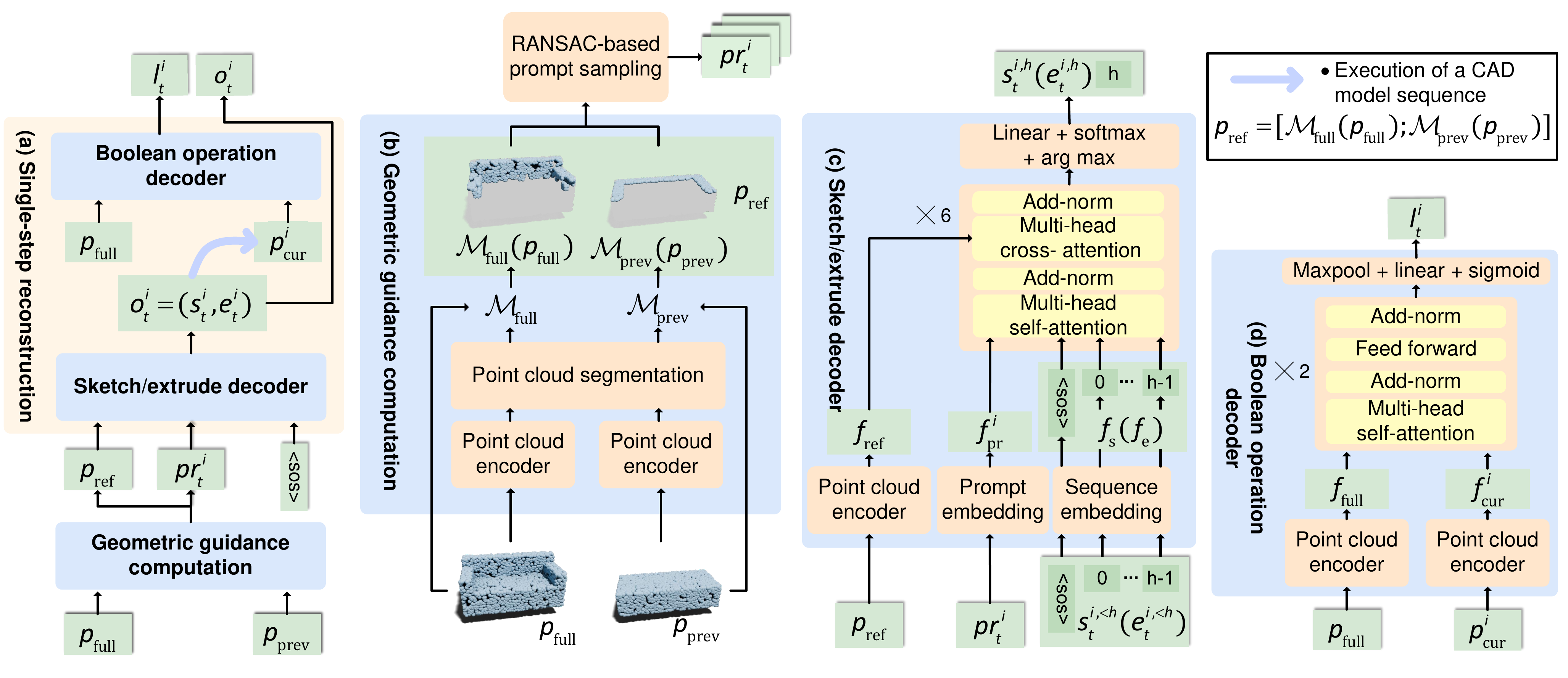}
    \caption{Illustration of \emph{Geometric Guidance Computation} and \emph{Single-Step Reconstruction}. Geometric guidance computation takes the point cloud $p_\text{full}$ and $p_\text{prev}$ as input. It outputs the difference between $p_\text{full}$ and $p_\text{prev}$ as $p_\text{ref}$ and a list of prompts $\text{pr}_t^i$ detected from $p_\text{ref}$. The single-step reconstruction module consists of a sketch/extrude decoder and a Boolean operation decoder.
    The prompt $pr_t^i$, the point cloud $p_\text{ref}$, and a start token <SOS> are fed into the sketch-extrude decoder to autoregressively predict a sketch $s_t^i$ and an extrude operation $e_t^i$ defining the extrusion cylinder $o_t^i=(s_t^i, e_t^i)$. The Boolean operation decoder predicts $l_t^i$ based on the input $p_\text{full}$ and $p_\text{cur}^i$, where $p_\text{cur}^i$ is obtained by executing $o_t^i$. At last, $l_t^i$ and $o_t^i$ are combined as the output of single-step reconstruction.}
    \label{fig:ssr}
\end{figure*}

    This section presents a module that takes the point cloud $p_\text{ref}$ and a single prompt $pr_t^i$ as input and outputs a candidate CAD modeling step $(o_t^i, l_t^i)$ as depicted in Fig.~\ref{fig:ssr}(a). This module is executed for each prompt $pr_t^i$ separately, but the point cloud $p_\text{ref}$ is shared between all separate executions of the module. 

    We first discuss the network architecture. We use three decoders (Fig.~\ref{fig:ssr}) to predict three components of a CAD modeling step: sketch decoder ($s$-dec) for $s_\text{t}$, extrude decoder ($e$-dec) for $e_\text{t}$, and Boolean operation decoder ($l$-dec) for $l_\text{t}$. In addition, we use a simple module for prompt embedding. See Fig.~\ref{fig:ssr} for an illustration of the architecture. We describe the individual components next.

    \head{Prompt embedding.} Prompt embedding is computed by embedding the prompt point cloud $pr_t^i$ using the point embedding part of the point cloud encoder. This embedding creates one token (feature vector) per point. We denote the output of this embedding as $f_\text{pr}^i$.

    \head{Sketch decoder and extrude decoder.}
    Both sketch decoder $s$-dec and extrude decoder $e$-dec adopt the same transformer architecture for autoregressive sequence prediction, implemented as $6$ stacked transformer decoder blocks shown in Fig.~\ref{fig:ssr}(c). Therefore, we describe them together.
    The input at each iteration of $s$-dec and $e$-dec are point cloud $p_\text{ref}$, prompt $pr_t^i$, and the current sequence of tokens ($s^{i,<h}_t$ or $e^{i,<h}_t$). Since both modules work in an auto-regressive manner, the modules themselves operate iteratively building the output sequences ($s^{i,<h}_t$ or $e^{i,<h}_t$) one step at a time. We use $h$ to denote the iteration. We use $s^{i,<h}_t$ as shorthand to denote the sequence $s^{i,0}_t, s^{i,1}_t, \ldots, s^{i,h-1}_t$.  In the first iteration, $s^{i,0}_t$ or $e^{i,0}_t$ are initialized with the start token $<SOS>$.
    First, all inputs are encoded: $p_\text{ref}$ is encoded with the point cloud encoder to yield $f_\text{ref}$, $pr_t^i$ is encoded to yield $f_\text{pr}^i$ and $s^{i,<h}_t$ ($e^{i,<h}_t$) is encoded to yield the sequence embedding of the sketch(extrude).
        
    The Queries of the self-attention and cross-attention layers in $s$-dec ($e$-dec) $q_{s-\text{dec}} (q_{e-\text{dec}}$) are the concatenation of $f_\text{pr}^i$ and the sequence embedding of $s^{i,<h}_t$ ($e^{i,<h}_t$). The same tokens also serve as keys and values for the self-attention layers. For the multi-head cross-attention layers, point cloud features $f_\text{ref}$, serve as Keys and Values.

    To output the next token, each token is assigned a softmax probability. An $\arg \max$ operator is followed to get the token value as follows:
    \begin{equation}
    \begin{split}
        s_{t}^{i,h} &= \mathop{\arg\max}\limits_{j} \Big (s\text{-dec}(q_{s-\text{dec}}^{<h}, \  f_\text{ref})^j \Big ), \ s\text{-dec}(\cdot, \cdot) \in \mathbb{R}^{1 \times \mathbf{Q}_s}, \\
        e_{t}^{i,h} &= \mathop{\arg\max}\limits_{j} \Big (e\text{-dec}(q_{e-\text{dec}}^{<h}, \  f_\text{ref})^j \Big ), \ e\text{-dec}(\cdot, \cdot) \in \mathbb{R}^{1 \times \mathbf{Q}_e},
    \end{split}
    \label{eq:dec-calculateion}
    \end{equation}
    $j$ is the component index of the output probability, with a range of $[0, \mathbf{Q}_s-1]$ in $s\text{-dec}(\cdot, \cdot)$ and $[0, \mathbf{Q}_e-1]$ in $e\text{-dec}(\cdot, \cdot)$. Q is the number of different discrete tokens for a sketch or an extrusion operation as defined in SkexGen.
    We then execute the predicted $s_t^i, e_t^i$ for a shape, from which we extract point clouds $p_\text{cur}^i$ and compute features $f_\text{cur}^i$.
    
    \head{Boolean operation decoder.}
    We implement $l$-dec as a transformer following a "Max-pooling + Linear + Sigmoid" layer, as depicted in Fig.~\ref{fig:ssr}(d).
    The $l$-dec takes $f_\text{cur}^i$ and $f_\text{full}$ as input, predicting a scalar in the range of $[0, 1]$ to represent $l_t^i$.
 
    \head{Training objective.}
    We train $s$-dec and $e$-dec using teacher forcing~\cite{kolen2001field} and cross-entropy loss $L_\text{CE}$, and train $l$-dec with binary cross-entropy loss $L_\text{BCE}$. 
    However, such an objective does not lead to a good performance. Experiments show the derived shape of the output sequence $o_\text{t}$ largely differs from the supervision-executed shape in geometric structure and appearance.
    We believe this is caused by the huge domain gap between sequential and geometric representation~\cite{ma2023multicad}, and we propose a geometric objective term to map the sequential output to the geometric representation space.
    Following Eq.~\ref{eq:geo-loss}, we construct a 3D bounding box of the executed shape of $o_\text{t}$, and calculate its intersection-over-union (IoU) with the bounding box of the supervision-executed shape,
    \begin{equation}
        L_\text{bbox} = - \ln \Big [ \text{IoU}\Big ( \text{bbox}(o_t^i), \text{bbox}(o_t^\text{gt}) \Big ) \Big]
    \label{eq:geo-loss}
    \end{equation}
    where $o_t^\text{gt}$ is the supervision of $o_\text{t}$, indicating the ground-truth extrusion cylinder for the step $t$. The bbox($\cdot$) is a differentiable operator to construct a 3D bounding box from $o_t^i$ or $o_t^\text{gt}$. The bounding box of an executed shape of a sketch-extrude sequence is illustrated in Fig.~\ref{fig:DSL}(c).
    %
    The overall objective function is Eq.~\ref{eq:overall-obj}, where $L_\text{bbox}$ is only enforced after $\text{ep}_\text{thres}$ training epochs.
    \begin{equation}
        L_o = L_\text{CE} + L_\text{BCE} + \alpha L_\text{bbox}, \quad     
        \alpha = \begin{cases} 
       \; 1,    & \mbox{if \, ep > ep}_\text{thres} \\
       \; 0,    & \mbox{otherwise}
    \end{cases}
    \label{eq:overall-obj}
    \end{equation}

    \head{Implementation details.}
    We leave implementation details of the differentiable bounding box construction bbox($\cdot$) in Supp.~\ref{subapp:ssr-geoloss}, and training strategy in Supp.~\ref{subapp:train}.

\section{Single-step Selection} \label{sec:sos}
    The single-step reconstruction module generates multiple candidate CAD modeling steps $(o_t^i, l_t^i)$. Here, we propose an architecture component that assigns probabilities to the different candidates.
    Though each choice $(o_t^i, l_t^i)$ consists of a sequence of tokens with corresponding probabilities (cf. Eq.~\ref{eq:dec-calculateion}), the probabilities are conditioned on a specific prompt. Therefore, we cannot directly compute the overall probability of each candidate CAD modeling step. 
    
    To rate different CAD modeling steps, we present a single-step selection module for addressing this task.
    The module is trained to measure the geometric fitness of candidate CAD modeling steps. which therefore avoids the confusion due to pertaining to the GT step, while being more efficient than brute-force enumeration that executes all candidate commands and compares them in retrospect. 
    We use the same mathematical notations as Sec.~\ref{sec:ssr} for convenience and readability.

    \begin{figure}
        \centering
        \includegraphics[width=\linewidth]{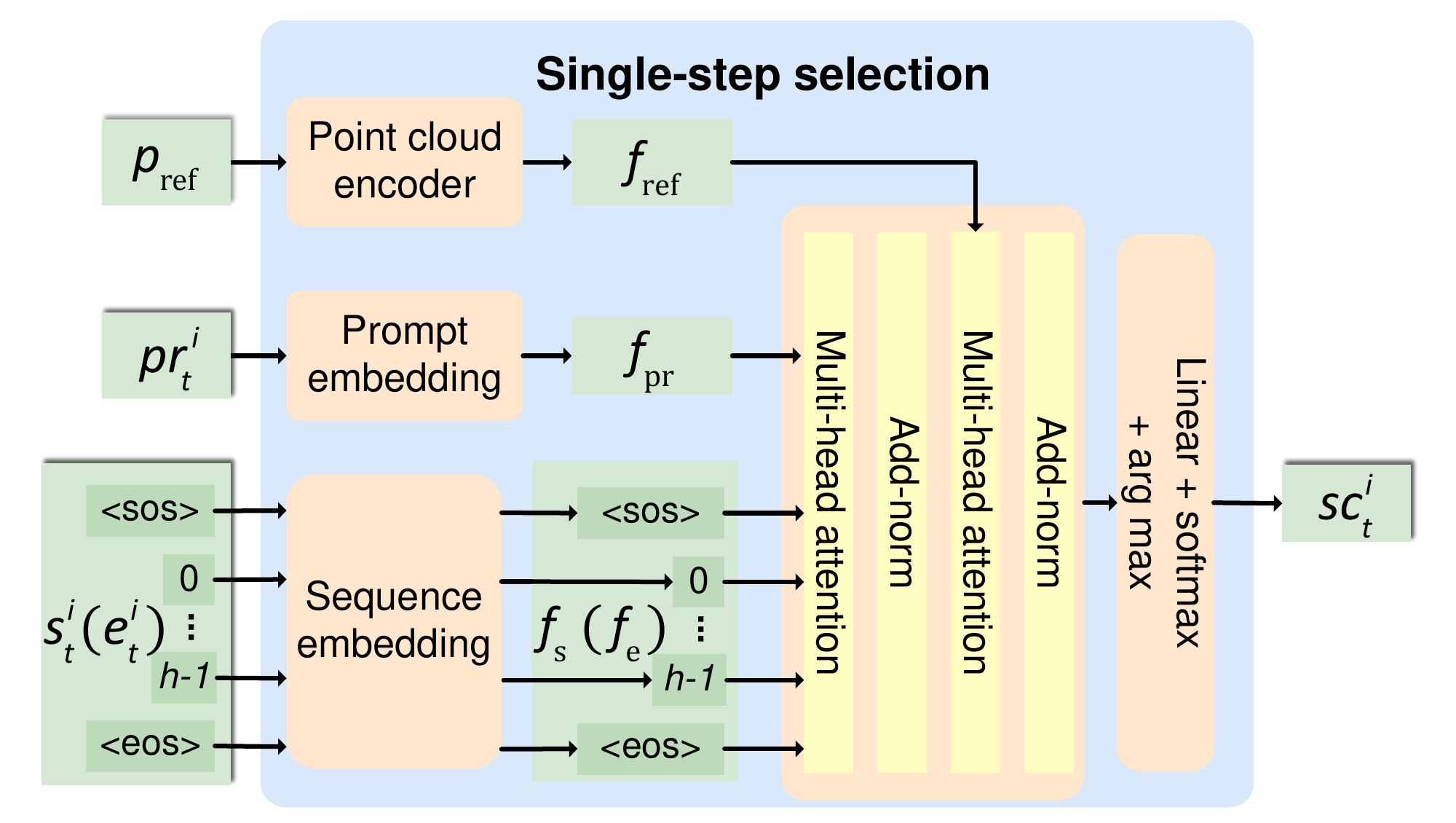}
        \caption{The architecture of the single-step selection module. At step $t$, the single-step selection module takes the point cloud $p_\text{ref}$, a prompt $pr_t^i$, and a set of candidate CAD modeling steps $\{(o_t^i, l_t^i)\}$, and predicts a fitness score $sc_t^i$ for each $(o_t^i, l_t^i)$ in the set.}
        \label{fig:SingleStepReconstruction}
    \end{figure}
    
    \head{Network architecture.} A visualization of the network architecture is shown in Fig.~\ref{fig:SingleStepReconstruction}. The single-step selection module (denoted as $f$-dec) takes the point cloud $p_\text{ref}$, a single prompt $pr_t^i$, and a candidate CAD modeling step $(o_t^i, l_t^i)$ as input and outputs a value $sc_\text{t}^i \in [0,1]$ representing the fitness of $(o_t^i, l_t^i)$.
    The $f$-dec is implemented in the same architecture as the sketch/extrude decoder in Fig.~\ref{fig:ssr}(c), except the output layer is modified from the "Linear + softmax + arg max" to the "Linear + sigmoid". For the input, the point cloud $p_\text{ref}$, the prompt $pr_t^i$, and $(o_t^i, l_t^i)$ are encoded to yield $f_\text{pr}$, $f_\text{pr}^i$ and the sequence embedding. The Query of $f$-dec ($q_{f-\text{dec}}$) is the concatenation of $f_\text{pr}^i$ and the sequence embedding of $(o_t^i, l_t^i)$. $f_\text{ref}$ serves as Key and Value. The output is as follows:
    \begin{equation}
        sc_t^i = f\text{-dec}(q_{f\text{-dec}}, \  f_\text{ref}), \ f\text{-dec}(\cdot, \cdot) \in \mathbb{R}^{1 \times 1}.
    \label{eq:f-dec}
    \end{equation}

    \head{Training objective.} The single-step selection module essentially performs a regression task and we choose the commonly used L1 loss as the objective function for step $t$, $L_t^\text{sc} = \big | sc_t^i - sc_t^{\text{gt}} \big |$.  
    $sc_t^{\text{gt}}$ denotes the supervision of the fitness score for $(o_t^i, l_t^i)$ as defined in Eq.~\ref{eq:sos-train-label-1}. Since $sc_t^{\text{gt}}$ cannot be directly obtained from the existing dataset~\cite{DBLP:conf/iccv/WuXZ21}, we calculate $sc_t^\text{gt}$ by comparing the geometric similarity between the executed shapes of $o_t^i$ and ground-truth $o_t^\text{gt}$. 
    The intuition is that $(o_t^i, l_t^i)$ cannot be concluded as unsuitable even if it is dissimilar to $(o_t^\text{gt}, l_t^\text{gt})$. Instead, it is more reliable to compare the similarity between their unique executions.
    We formulate $sc_t^\text{gt}$ as the bounding box intersection-over-union (IoU) between the executed shape of $o_t^i$ and $o_t^\text{gt}$, multiplied by the binary value of the $l_t^i$ and $l_t^\text{gt}$ consistency, where the $\text{bbox}(\cdot)$ is the same as that in Eq.~\ref{eq:geo-loss}.
    
    \begin{equation}
        sc_t^\text{gt} = 
        \begin{cases} 
       \text{IoU} \Big (\text{bbox}(o_t^i), \text{bbox}(o_t^\text{gt}) \Big ),    & \text{if \ } l_t^i = l_t^\text{gt}, \\
       \; 0,    & \text{otherwise}.
        \end{cases}
    \label{eq:sos-train-label-1}
    \end{equation}
     
    We use the bounding box IoU to approximate the geometric similarity for two reasons.
    First, calculating the bounding box IoU is more straightforward compared to calculating the fine-grained IoU between shapes. 
    Second, the bounding box can accurately reflect the principal parameters of the sketch-extrude sequence constructing the shape (cf. Extrude value, Sketch plane orientation, etc. in Fig.~\ref{fig:DSL}).

\section{Experiments and Analysis}
\label{sec:expr}
    In this section, we provide a quantitative and qualitative evaluation, an ablation study, and a robustness test of our method.

\subsection{Experimental Setup}
    \head{Datasets.} We use the DeepCAD~\cite{DBLP:conf/iccv/WuXZ21} dataset for training. Following the dataset split of DeepCAD, we train the networks on the DeepCAD train set and report results on the DeepCAD test set. We also report the cross-dataset generalization based on the Fusion360~\cite{DBLP:journals/tog/WillisPLCDLSM21} dataset. For all input CAD models, we randomly sample 8192 points on the surface as input point clouds.
    
    \head{Evaluation metrics.} We evaluate the geometric quality following \cite{DBLP:journals/tog/GuoLPG22} and sequence fidelity using NLL used to evaluate similar auto-regressive generative models. Geometric metrics are calculated between the input CAD models and the execution of predicted CAD modeling sequence $O$ with the following terms:
    \begin{itemize}[noitemsep, leftmargin=5mm]
    \item Chamfer distance (CD) and Hausdorff distance (HD) of 16382 randomly sampled points on the geometry surface of two shapes for geometry distance. 
    \item Edge chamfer distance (ECD) to compute the distance between edge points of two CAD models to reflect the structure similarity. 
    \item Normal consistency (NC) to measure the cosine similarity between corresponding normals as the smoothness of predicted shapes. 
    \item Invalidity Ratio (IR) to measure the ratio of CAD modeling sequences that cannot be executed.
    \end{itemize}
    We use the point cloud of the ground-truth CAD model and the point cloud of output CAD modeling sequence execution to compute geometric metrics. For a scale-invariance comparison, we normalize the ground-truth CAD model and reconstruction into a unit bounding box keeping the aspect ratio. 
    To determine whether a reconstruction is invalid, we retain the output of each step ($O_0, O_1,\ldots$). If none of them leads to a successful execution, we consider it to be invalid. 
    We use the Negative log-likelihood (NLL) of the auto-regressive model to evaluate the sequence fidelity produced by the single-step reconstruction module. Specifically, given the point cloud of the input CAD model and that of the ground-truth previous state, we predict the logits of the next CAD modeling step and calculate the NLL. We evaluate the quality of sketch and extruding, separately with \emph{ske\_NLL} and \emph{ext\_NLL}.

    \head{Baselines.} We compare with four SoTA methods~\cite{DBLP:conf/iccv/WuXZ21,DBLP:conf/icml/XuJLWF23,DBLP:conf/cvpr/UyCSGLBG22,DBLP:conf/cvpr/LiG0Y23,DBLP:conf/cvpr/XuPCWR21} focusing on the reverse engineering of CAD modeling sequences. 
    We compare with these methods by running the available codes and pre-trained weights with the same test splits for a fair comparison.
    Point2Cyl~\cite{DBLP:conf/cvpr/UyCSGLBG22} applies instance segmentation on point cloud of the input CAD model. Each instance is treated as a potential extrusion cylinder and another segmentation is applied to locate the cross-section (base) and side surface (barrel). Afterward, Point2Cyl fit for an extrusion operation and a sketch utilizing base and barrel.
    SECAD-Net~\cite{DBLP:conf/cvpr/LiG0Y23} requires both voxel and point cloud as inputs, We apply mesh-voxelization~\cite{Stutz2018CVPR, Stutz2017} to convert the mesh of a CAD model into a $64 \times 64 \times 64$ voxel. It outputs a fixed number of CAD modeling steps and we follow its official implementation to output 4 CAD modeling steps.
    Since SECAD-Net employs unsupervised learning, it can overfit to each test data to get the best reconstruction result.
    HNC-CAD is initially designed for CAD modeling sequence generation. We train HNC-CAD by feeding it with the point cloud features of the input CAD model, without changing any output to adapt it to perform CAD modeling sequence reconstruction.
    We do not compare with~\cite{DBLP:conf/siggrapha/LambourneWJZSM22} and the concurrent work~\cite{khan2024cadsignet}, because they do not release the source code.

    \head{Implementation details.}
    For geometric guidance computation (Sec.~\ref{sec:cpg}), we train the point cloud segmentation network for 80 epochs with a batch size of 256. An AdamW optimizer with an initial learning rate of 1e-3 is employed for optimization and the learning rate is scheduled by a cosine annealing scheduler until the minimal learning rate of 1e-6.
    We train the single-step reconstruction (Sec.~\ref{sec:ssr}) based on the objective function of Eq.~\ref{eq:overall-obj}.
    The network is trained for 200 epochs with a batch size of 224 on 1 A100-80G GPU, using the AdamW optimizer with a learning rate of 1e-3.
    The learning rate drops to 9e-5 after the initial 90 epochs and gradually decreases to the minimal learning rate of 5e-6 based on the cosine annealing scheduler. We empirically set the $\text{ep}_\text{thres}$ in Eq.~\ref{eq:overall-obj} as 30.
    The single-step selection (sec.~\ref{sec:sos}) is trained for 100 epochs with a batch size of 224 on 1 A100-80G GPU, using the same optimizer and learning rate scheduler as the single-step reconstruction.

\subsection{Comparison with the State of the Art}

\begin{table}[t]
    \centering
    \caption{Quantitative comparisons of geometric metrics of different methods on DeepCAD and Fusion360. CD, HD, and ECD are scaled by $10^2$. In DeepCAD, our method improves over SECAD-Net by $10\%$ on geometric error (CD, HD) and $16\%$ on structural error (ECD). In Fusion360, our method also shows better geometric quality than SECAD-Net, but has a lower invalid reconstruction ratio (IR).}
    \setlength{\tabcolsep}{1mm}
    \resizebox{\linewidth}{!}{
    \begin{tabular}{c|c||ccccc}
    \toprule
    Dataset &Method       &CD $\downarrow$ & HD $\downarrow$ & ECD $\downarrow$  &  NC $\uparrow$  & IR ($\%$) $\downarrow$\\ \midrule
    &DeepCAD        &4.25 & 39.25 & 19.33 & 0.49 & 7.14\\
    &HNC-CAD        &1.09 & 20.23 & 5.94 & 0.75 & 0.32\\
    DeepCAD &Point2cyl      &1.00 & 20.93 & 20.45  & 0.73 & \textbf{0.0} \\
    &SECAD-Net      &0.42 & 9.96 & 5.54  & 0.73  & 0.38 \\ 
    &Ours (PS-CAD)           &\textbf{0.21} & \textbf{8.66} & \textbf{4.65}  & \textbf{0.89} & 0.43 \\ \midrule
    &HNC-CAD        & 1.38 & 20.61 & 9.24 & 0.52 & \textbf{0.33} \\
    Fusion360 &SECAD-Net      & 0.69 & 12.76 & 5.15 & 0.64 & 3.82 \\ 
    &Ours (PS-CAD)  & \textbf{0.57} & \textbf{12.0} & \textbf{5.02} & \textbf{0.67} & 1.51   \\ \bottomrule
    \end{tabular}
    }
    \label{tab:comparison_pcd}
\end{table}

\begin{table}[t]
    \centering
    \caption{Quantitative comparison of the negative log-likelihood (NLL) of sequentially-based reconstruction methods.}
    \begin{tabular}{c|cc}
    \toprule
         Method & ext\_NLL $\downarrow$ & ske\_NLL $\downarrow$ \\ \midrule
         HNC-CAD & 38.66 & 103.33 \\ 
         Ours (PS-CAD) & \textbf{1.61} & \textbf{25.51} \\ \bottomrule
    \end{tabular}
    \label{tab:NLL_comp}
\end{table}

\begin{figure*}
    \centering
    \includegraphics[width=\textwidth]{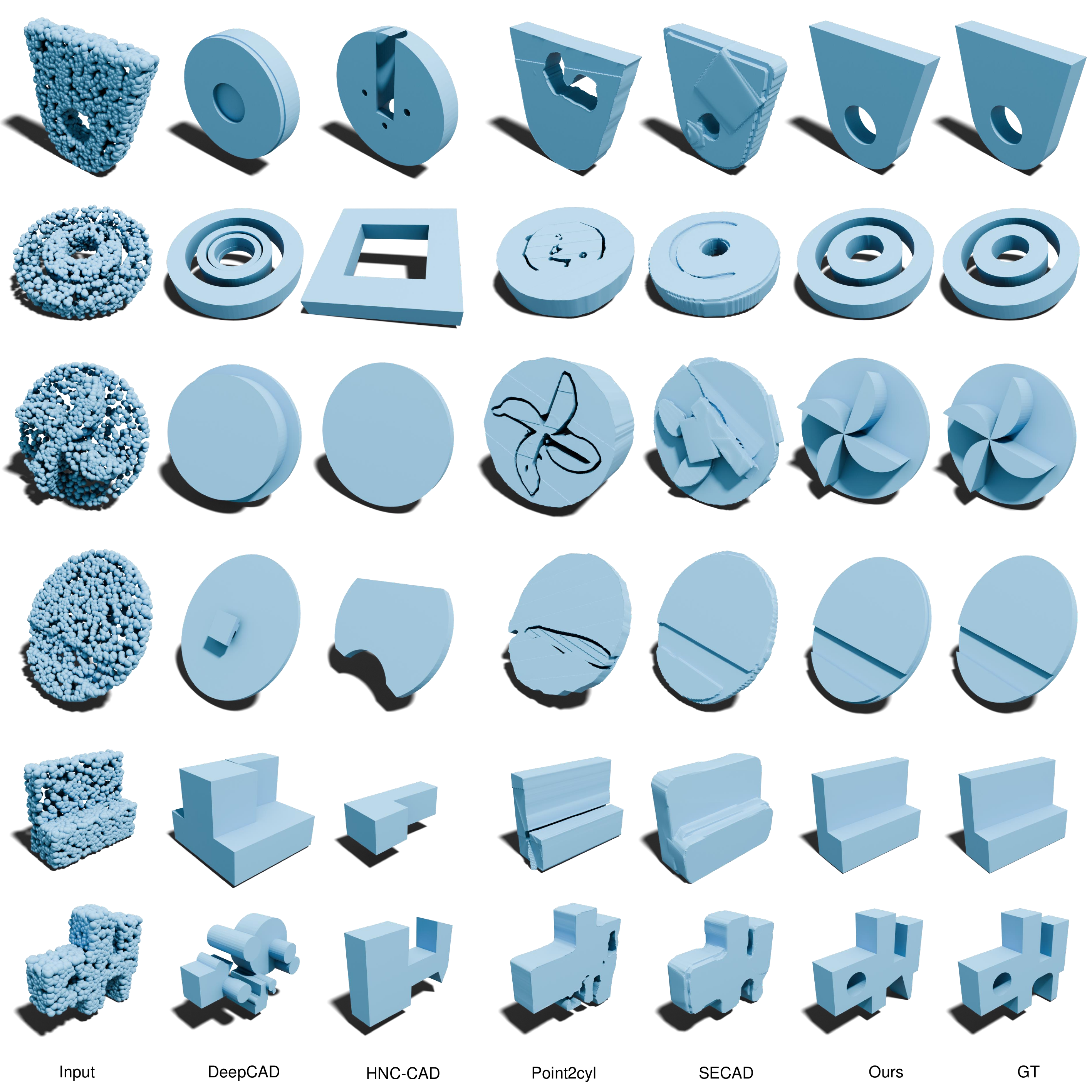}
    \caption{Qualitative comparison of different methods on DeepCAD. Please zoom in to see the deviations from ground truth. }
    \label{fig:comparison-DeepCAD}
\end{figure*}

\begin{table*}[t]
    \centering
    \caption{Ablation study of the single-step reconstruction and the single-step selection module. For single-step reconstruction, we measure the effect of the different inputs when reconstructing one CAD modeling step. $p_\text{full}$: point cloud of the input CAD model; $[p_\text{full};p_\text{prev}]$: point cloud of the input CAD model is concatenated with point cloud of previous state execution; $p_\text{ref}$: the distinct regions between $p_\text{full}$ and $p_\text{prev}$. $pr$: the geometric prompt. We also evaluate the proposed bounding box loss  $L_\text{bbox}$ in Eq.~\ref{eq:geo-loss}. For the single-step selection module, we compare the performance of different selection strategies. "geo": the greedy selection based on the geometric criteria implemented as chamfer distance; "heur":  selecting the candidate extrusion cylinder with the largest volume; "rand": random selection; "ours": using the single-step selection module.}
    \setlength{\tabcolsep}{2mm}
    \resizebox{0.8\textwidth}{!}{
    \begin{tabular}[width=\textwidth]{ccccc|cccc||cccc}
    \toprule
    \multicolumn{5}{c|}{Single-step reconstruction}& \multicolumn{4}{c||}{Single-step selection}& \multicolumn{4}{c}{Metrics} \\ \midrule
    $p_\text{full}$ & $[p_\text{full};p_\text{prev}]$& $p_\text{ref}$& \ $pr$& $L_\text{bbox}$& geo &heur& rand& ours& CD& ext\_NLL& ske\_NLL & IR ($\%$) \\ \midrule
    $\surd$& & & & & $\surd$ & & &                             & 1.21 & 5.77 & 132.85& 7.53\\
    & $\surd$& & & & $\surd$ & & &                      & 0.67 & 2.24 & 91.03 &6.85\\
     & & $\surd$& & & $\surd$& & &                & 0.37 & 1.68 & 33.45 & 6.99\\
     & & $\surd$& $\surd$& & $\surd$ & & &        & 0.61 & 1.98 & 47.84 & 1.37\\
     & & $\surd$& $\surd$& $\surd$& $\surd$ & & & & 0.28 & 1.59 & 30.10 & 0.69\\ \midrule
     & & $\surd$& $\surd$& $\surd$& & $\surd$ & & & 0.95 & 2.49 & 100.08 & 1.37 \\ 
     & & $\surd$& $\surd$& $\surd$& & & $\surd$ & & 1.48 & 7.95 & 213.27 & 1.51 \\ 
     & & $\surd$& $\surd$& $\surd$& & & & $\surd$ & 0.19 & 1.49 & 19.63    & 0.41\\ \bottomrule 
    \end{tabular}
    }
    \label{tab:ablation-ssr}
\end{table*}

    \head{Main comparison.}
    In Tab.~\ref{tab:comparison_pcd}, we quantitatively show the performance of all the competing methods.
    Directly predicting a CAD modeling sequence as the CAD DSL sequence, e.g., DeepCAD~\cite{DBLP:conf/iccv/WuXZ21} and HNC-CAD~\cite{DBLP:conf/icml/XuJLWF23} for this task results in poor performance because aligning point clouds with a CAD DSL is difficult. 
    Our proposed method PS-CAD significantly improves upon all SOTA methods on all geometric metrics on DeepCAD. We can reduce the geometry errors (CD and HD) by at least $10\%$, and the structural error (ECD metric) by about $15\%$. Moreover, the proposed method can attain the best normal quality.
    The generalization comparison on Fusion360 yields better results than HNC-CAD and SECAD-Net. We reduce the geometry errors by at least $5\%$ and the structural error by about $2.5\%$.
    
    We show a comparison of the NLL in Tab.~\ref{tab:NLL_comp}. We can only compare to the similar auto-regressive model HNC-CAD and we note a huge improvement in the metrics. This suggests that geometric guidance can greatly improve the accuracy of predicting CAD DSL sequences.

\begin{figure}
    \centering
    \includegraphics[width=\linewidth]{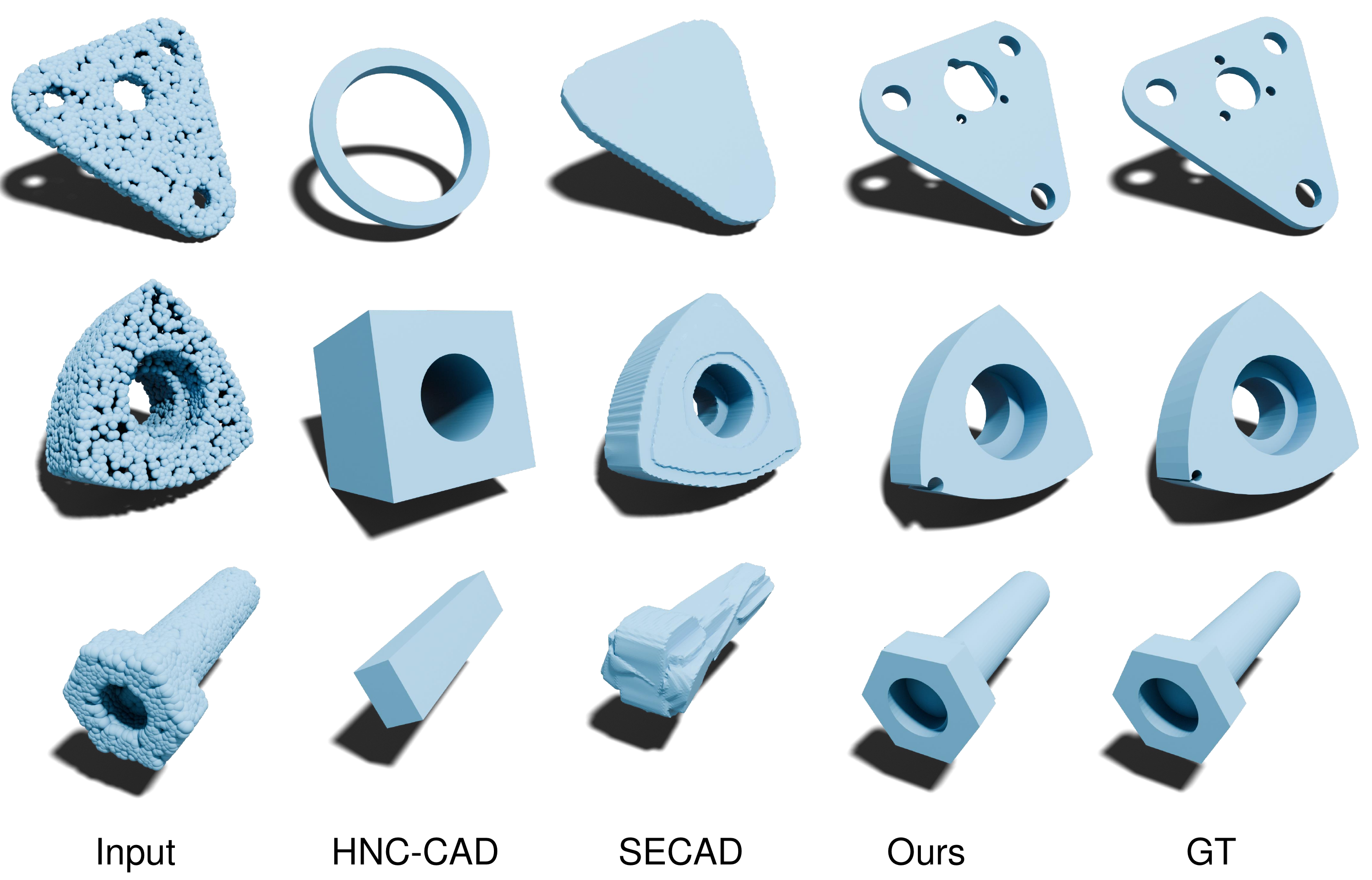}
    \caption{Qualitative comparison of different methods on Fusion-360 to evaluate generalization capabilities. None of the methods were trained on Fusion-360.}
    \label{fig:comparison-Fusion}
\end{figure}

    \head{Qualitative comparison.}
    In Fig.~\ref{fig:comparison-DeepCAD} and Fig~\ref{fig:comparison-Fusion}, we qualitatively show the performance on DeepCAD and Fusion360 of the competing methods. Note that our point cloud visualization shows the overall structure well, but seeing some of the detailed structures in the input point cloud would require inspecting the point cloud in an interactive editor from multiple views (See Appendix Fig.~\ref{fig:bird-eye view pcd}).
    DeepCAD and HNC-CAD output a CAD DSL sequence. They can correctly reconstruct the coarse appearance of the input CAD models, but the structure as well as the details are often wrong. (see column DeepCAD, HNC-CAD in Fig.~\ref{fig:comparison-DeepCAD}, column HNC-CAD in Fig.~\ref{fig:comparison-Fusion}). 

    Point2Cyl and SECAD-Net reconstruct sketches with implicit representations and generally perform better than DeepCAD and HNC-CAD. Still, Point2Cyl and SECAD-Net also have problems with the details as well as the structure.  Point2Cyl often predicts the main shape correctly, but it adds incorrect and overly complicated details  (all rows in Fig.~\ref{fig:comparison-DeepCAD}). SECAD-Net sometimes does not preserve the basic structures, e.g. crevices in the input (see row 2,3,6 in Fig.~\ref{fig:comparison-DeepCAD} and rows in Fig.~\ref{fig:comparison-Fusion}) or invents erroneous details, e.g., a square extrusion cylinder in row 1 in Fig.~\ref{fig:comparison-DeepCAD}.
    By contrast, our method reproduces the structure and the details of the input better.
    Our method is also capable of reconstructing detailed geometry of inputs from a cross-domain dataset Fusion360, e.g., different kinds of holes in Fig.~\ref{fig:comparison-Fusion}.

\subsection{Ablation Studies}

\begin{figure*}
    \centering
    \includegraphics[width=\textwidth]{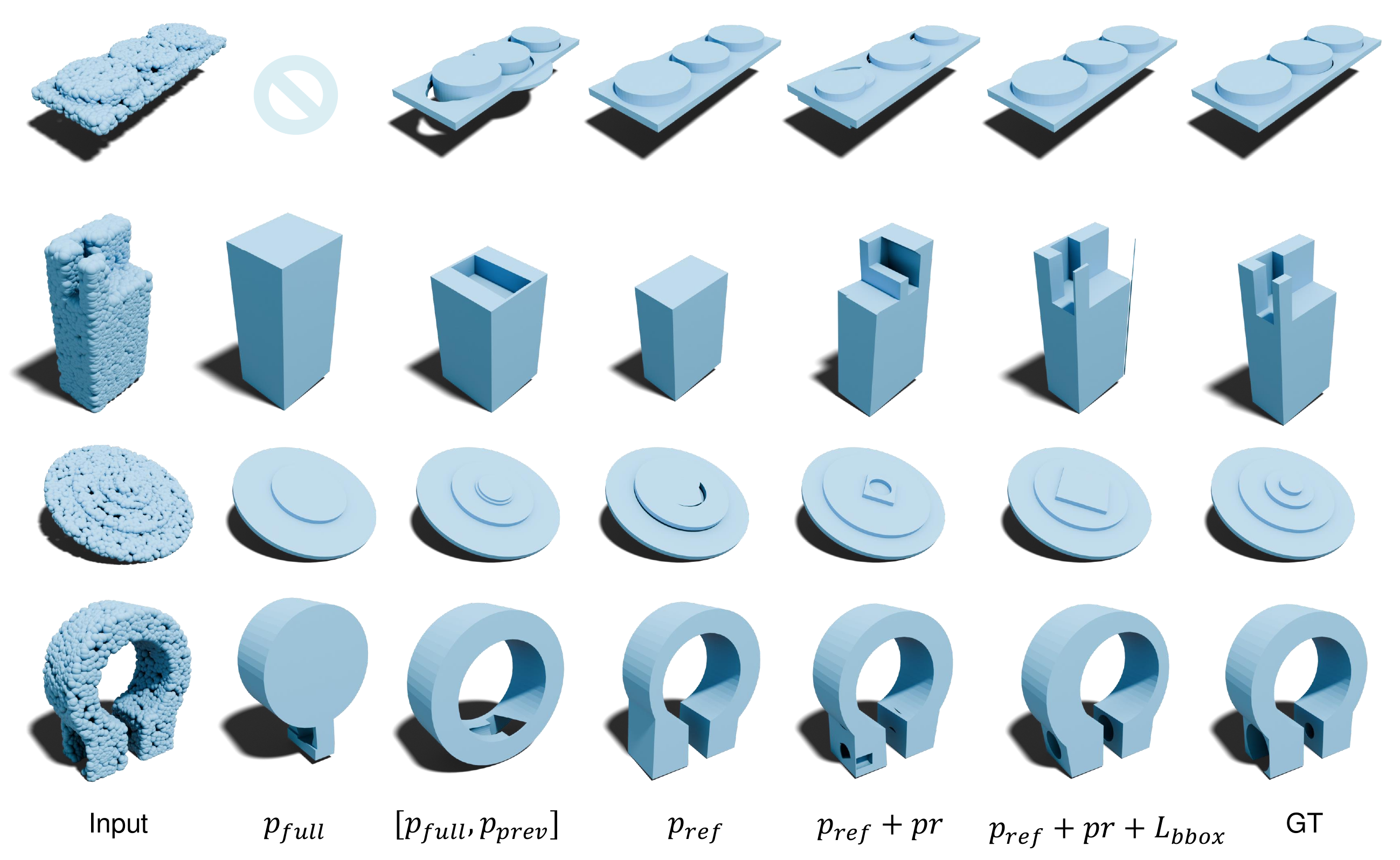}
    \caption{Ablating different designs in single-step reconstruction. See the text for details about $p_\text{full}$, $[p_\text{full};p_\text{prev}]$, $p_\text{ref}$, $pr$, $L_\text{bbox}$.}
    \label{fig:ablation-reconstruction}
\end{figure*}

    In Tab.~\ref{tab:ablation-ssr}, we first show a set of ablations on the designs of the single-step reconstruction. We randomly select 730 input models for the ablation studies, which accounts for approximately 10\% of the total test dataset. Note that we evaluate the designs of the single-step reconstruction with the selection strategy "geo". This is because the single-step selection is based on the output of the single-step reconstruction. When the output quality varies due to the ablation setting, the rating result of the single-step selection introduces extraneous factors that affect the ablation.
    
    \head{The effect of $p_\text{ref}$.} The first three rows of Tab.~\ref{tab:ablation-ssr} ablate the input fed to single-step reconstruction. Only given $p_\text{full}$ leads to poor performance in geometric quality and sequence fidelity. It can only reconstruct the main structure and loses many details. Sometimes an invalid reconstruction is produced (see Fig.~\ref{fig:ablation-reconstruction} $p_\text{full}$).  With the help of $p_\text{prev}$, there is a noticeable reduction in the extrude NLL (ext\_NLL). By explicitly identifying the distinct regions of $p_\text{full}$ and $p_\text{prev}$ with a binary mask $\mathcal{M}$, the network can distinguish which part has already been completed for reconstruction, and which part still needs further work. Using the mask gives a cleaner geometric context and improves the sequence fidelity, reflected in further reductions in sketch NLL and extrude NLL.

    \head{The effect of $pr$.} By providing a prompt to the single-step reconstruction, albeit with a decrease in geometric reconstruction quality, it improves the reconstruction invalid rate. This suggests that the prompt improves the tolerance to input CAD models with different shapes. We believe the source of this ability is that the prompt can provide guidance for local geometric context encoding. This local geometric context, compared with $p_\text{ref}$, can support the reconstruction of one valid CAD modeling step with better efficiency. 

\begin{figure*}
    \centering
    \includegraphics[width=\linewidth]{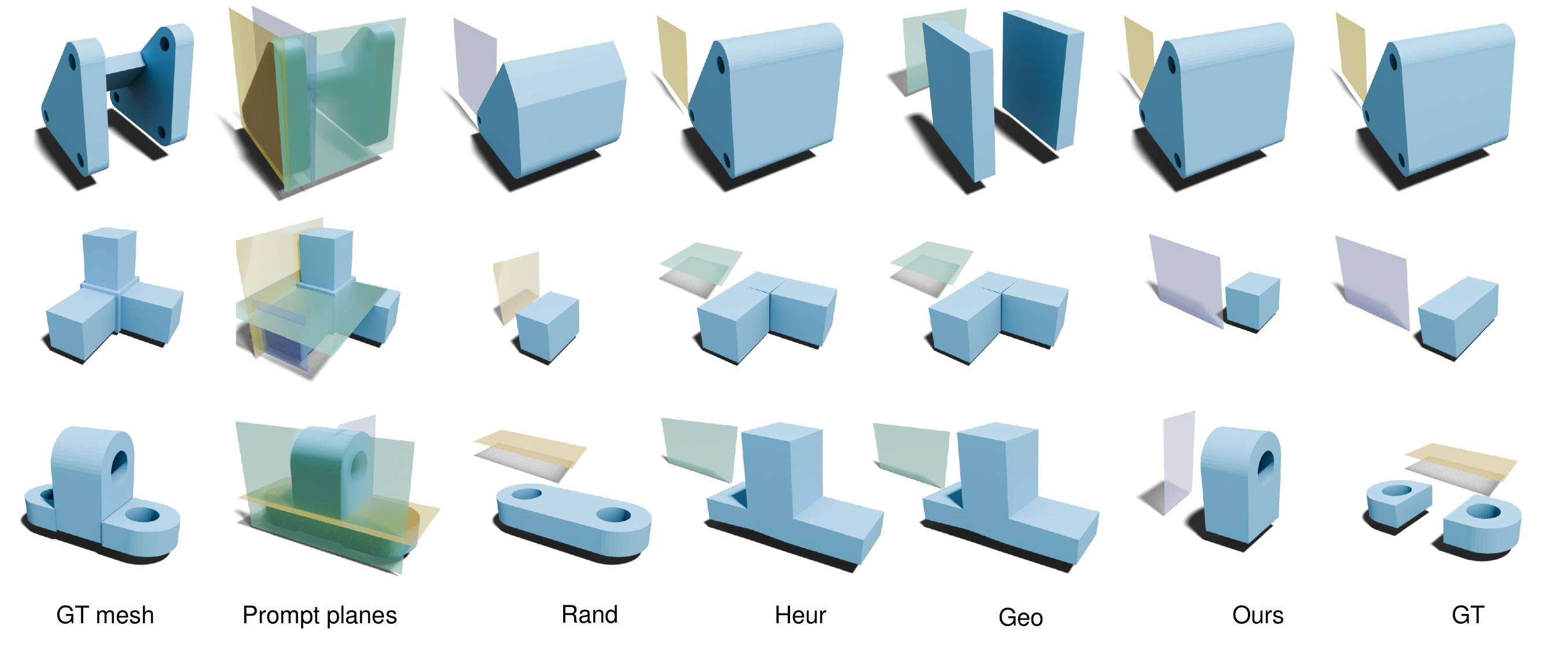}
    \caption{Comparisons of different prompt selection methods. See the paper text for an explanation of the different selection methods.}
    \label{fig:ablation-selection}
\end{figure*}

    \head{Single-step selection.}
    Afterward, we fix the setting of single-step reconstruction and validate the single-step selection by comparing it with different selection strategies. Among the other selection strategies, "geo" is based on the geometric quality of the candidate CAD modeling step. Take the step-$t$ for example, we make temporary step-$t$  $O_t^i$ by combining the candidate CAD modeling step $(o_t^i, l_t^i)$ with the state-$t-1$ output $O_{t-1}$. The execution of $O_t^i$ as point cloud is compared with point cloud $p_\text{full}$ of the input CAD model to calculate the chamfer distance. "geo" selects the $O_t^{i*}$ with the lowest chamfer distance as the step-$t$ output.
    "heur" is to select the extrusion cylinder $o_t^{i-}$ with the largest volume. It follows a design convention that a CAD model is usually created by first constructing the main structure and then designing the details.
    "rand" is to randomly select a candidate CAD modeling step.
    The results show that the heuristic and geometric-based selection do not work as well for all inputs because they do not consider the structure of the input CAD model. For example in Fig.~\ref{fig:ablation-selection} row 3, the input is a combination of two extrusion cylinders. However, the side surface is sampled as a prompt and leads to a candidate CAD modeling step with a "T" shaped extrusion cylinder. This "T-shape" extrusion cylinder has the lowest chamfer distance with the input and has the largest volume among all the candidates. In Fig.~\ref{fig:ablation-selection} row 1, the heuristic-based selection works well, but the geometric-based selection performs badly and creates two boxes using the green prompt. The reason is that the chamfer distance of the CAD modeling step stemming from the green prompt is slightly better. The geometric-based selection then chooses the wrong candidate. By contrast, the single-step selection performs well in both cases, while it may select a different candidate from the ground-truth, the selected candidate can still correctly reflect the input structure.

\subsection{Robustness test}
    In this section, we test the robustness of the single-step reconstruction module with respect to the prompt. We specifically test the effect of prompt sampling under different conditions, including the number of points of a sampled prompt, the source (sampled from either  $p_{full}$ or $p_{ref}$), and the quality of the prompt-sampling plane. Finally, we visualize some reconstruction results when our method is fed with noisy point clouds of the input CAD model.
    
    \head{Prompt.} In Tab.~\ref{tab:pr-1}-Sampling location, we detect planes from point clouds of different sources ($p_{full}$ or $p_{ref}$) and sample prompts from these planes. Results show that if the planes are sampled from the point cloud $p_\text{full}$ of the input CAD model in step-wise reconstruction, the performance decreases in all metrics. The extrude and sketch NLL deteriorate more than $100\%$ (0.19$\rightarrow$0.48) and $400\%$ (1.49$\rightarrow$3.18). If a prompt is always from $p_\text{full}$, it lacks the ability to adapt to the dynamically changing geometric context in the iterative reconstruction.

    In Tab.~\ref{tab:pr-1}-Sampling number, we verify the effect of the number of points within a prompt. Given planes detected from the point cloud $p_\text{ref}$, the prompts with fewer points lead to a higher reconstruction invalid ratio (prompt points 64$\rightarrow$16, IR 0.41$\rightarrow$0.83).
    
    Since the prompt relies on RANSAC-detected planes, they do not always match the surface of extrusion cylinders to a high degree.  In Tab.~\ref{tab:pr-2}-Perturbations on planes, we sample prompts from planes with different corruptions, including plane deficiency or (and) Gaussian noise. Results show that our method is robust to the Gaussian noise added on prompts. When a prompt is sampled from a plane with deficiency, it cannot capture the sketch information and produces weaker reconstructions (ske\_NLL from 19.51 to 33.99). This also leads to a higher invalid reconstruction ratio (IR) and chamfer distance (CD).

     It is also critical to validate the effect of the RANSAC hyperparameters on prompt sampling. We test two of the hyperparameters in Tab.~\ref{tab:pr-2}-Hyperparameters of RANSAC, namely t and d. t is the threshold value to determine when a point is detected to fit a plane. If t increases, the detected plane deviates from the real sketch plane and produces worse sketches (ske\_NLL from 19.51 to 81.64, when t from 1e-3 to 1e-2). Due to the association of extrusion and the sketch plane, the extrude quality also decreases (ext\_NLL 1.49 $\rightarrow$ 1.95). These factors result in higher IR and CD. d is the number of inlier points required to assert that a plane is fitted. If d is provided with a larger number (64 $\rightarrow$ 128 $\rightarrow$ 256), some sketches located on a tinner plane would be omitted and this results in a higher IR (0.27 $\rightarrow$ 0.41 $\rightarrow$ 2.30).

\begin{table}[]
    \centering
    \caption{Effect of prompts sampled under different settings. In Sampling location, $p_\text{full}$ denotes prompts sampled from the point cloud of the input CAD model to reconstruct an arbitrary CAD modeling step. $p_\text{ref}$ denotes prompts from point cloud $p_\text{ref}$. In Sampling number, prompts with different numbers of points are evaluated (16, 32, and 64 respectively).}
    \begin{tabular}{cc|ccc||cccc}
    \toprule
    \multicolumn{2}{c|}{\makecell{Sampling \\ location}} & \multicolumn{3}{c||}{\makecell{Sampling \\ number}} & \multicolumn{4}{c}{Metrics} \\ \midrule
    $p_\text{full}$ & $p_\text{ref}$ & 16  & 32 & 64 & CD & ext\_NLL & ske\_NLL & IR \\ \midrule
    $\surd$& & & & $\surd$      & 0.48 & 3.18 & 107.04& 1.78\\
    &$\surd$ & & & $\surd$      & 0.19 & 1.49 & 19.63 & 0.41\\ \midrule
    & $\surd$&$\surd$& &        & 0.19 & 1.53 & 20.46 & 0.83\\ 
    & $\surd$& &$\surd$ &       & 0.20 & 1.51 & 20.53 & 0.54\\ \bottomrule
    \end{tabular}
    \label{tab:pr-1}
\end{table}

\begin{table}[t]
    \centering
    \caption{Effect of prompts sampled from low-quality planar surfaces. In Perturbations on planes, Gaussian is to add Gaussian noise on the detected planes; deficiency 50\% means randomly covering half of the detected plane. In Hyperparameters of RANSAC, t is the threshold value to determine when a point is detected to fit for a plane, and d is the number of inlier points required to assert that a plane is fitted.}
    \setlength{\tabcolsep}{1mm}
    \resizebox{\linewidth}{!}{
    \begin{tabular}{cc|cc||cccc}
    \toprule
    \multicolumn{2}{c|}{\makecell{Perturbations \\ on planes}} & \multicolumn{2}{c||}{\makecell{Hyperparameters \\ of RANSAC}} & \multicolumn{4}{c}{Metrics} \\ \midrule
    Gaussian & deficiency 50\% & t & d & CD & ext\_NLL & ske\_NLL & IR \\ \midrule
    &        & 1e-3 & 128               & 0.19 & 1.49 & 19.51 & 0.41 \\
    $\surd$ && 1e-3 & 128               & 0.19 & 1.53 & 20.33 & 0.54 \\
    &$\surd$ & 1e-3 & 128               & 0.34 & 1.69 & 33.99 & 1.65 \\
    $\surd$ &$\surd$ & 1e-3 & 128       & 0.35 & 1.70 & 34.10 & 1.78 \\ \midrule
    & & 1e-3 & 64                       & 0.23 & 1.62 & 18.65 & 0.27 \\
    & & 1e-3 & 256                      & 0.23 & 1.65 & 21.14 & 2.30 \\
    & & 3e-3 & 128                      & 0.25 & 1.60 & 20.62 & 0.27 \\ 
    & & 5e-3 & 128                      & 0.37 & 1.83 & 46.70 & 2.33 \\ 
    & & 1e-2 & 128                      & 0.39 & 1.95 & 81.63 & 1.92 \\ \bottomrule
    \end{tabular}
    }
    \label{tab:pr-2}
\end{table}

\begin{figure*}
    \centering
    \includegraphics[width=\textwidth]{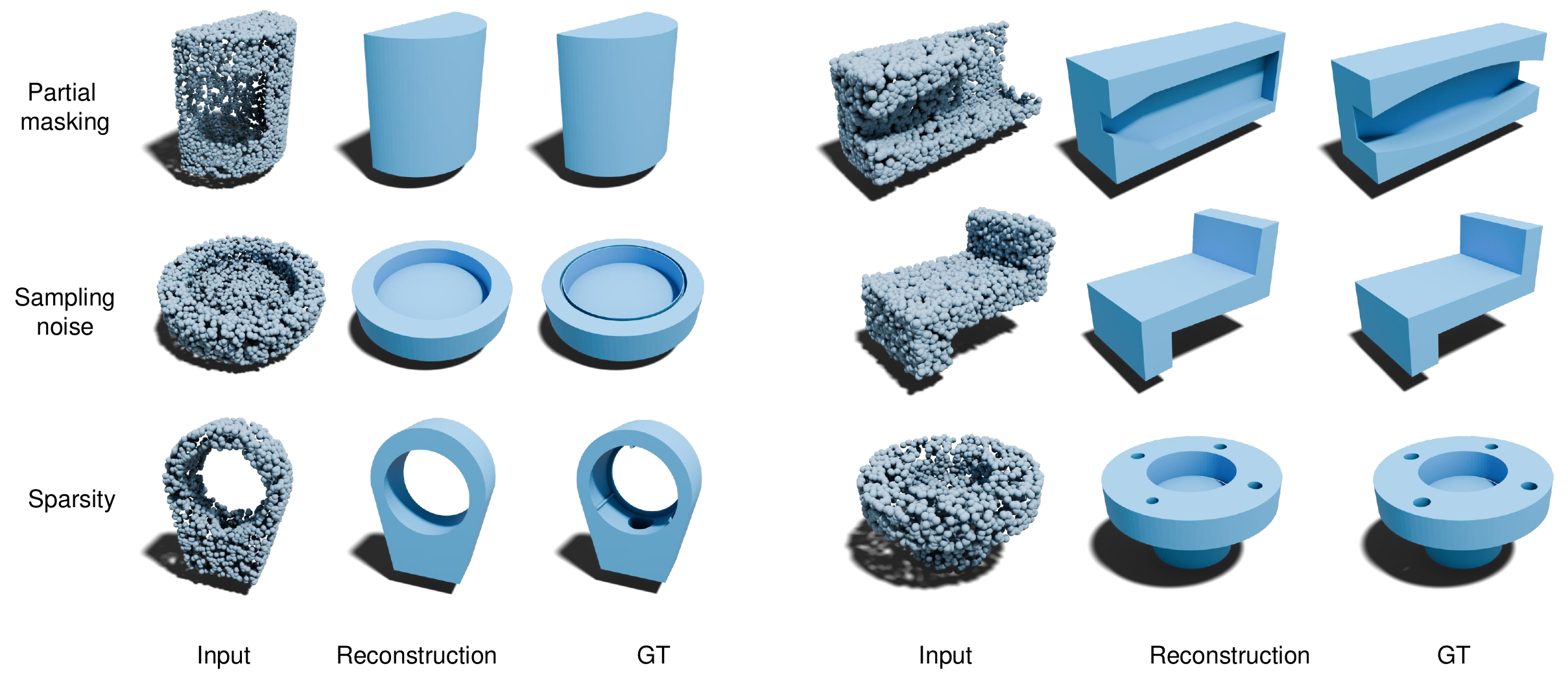}
    \caption{Robustness validations for different cases, including partial point masking, adding noises, and increasing point cloud sparsity.}
    \label{fig:robust}
\end{figure*}

\begin{figure*}
    \centering
    \includegraphics[width=\linewidth]{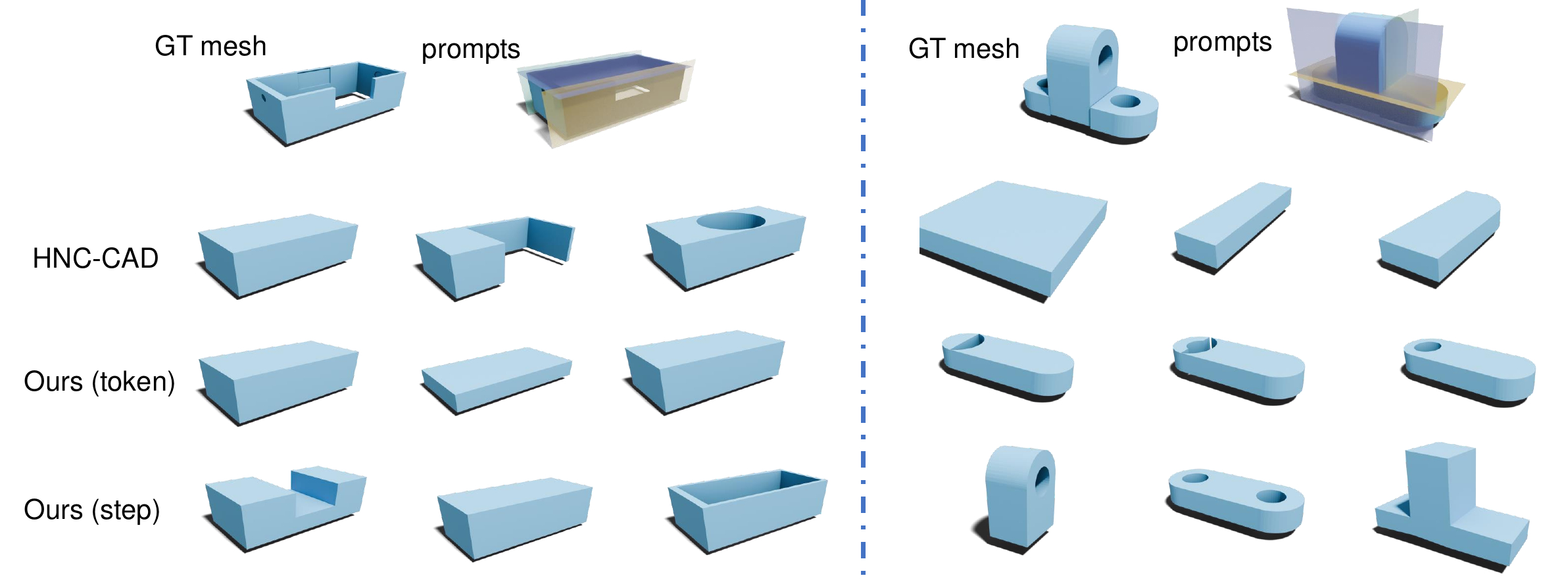}
    \caption{Comparisons of two token-based sampling methods and ours on the sampling variations.}
    \label{fig:ablation-sample-comp}
\end{figure*}

\begin{table}[t]
    \centering
    \caption{Evaluation of different numbers of input points and point cloud encoder layers.}
    \begin{tabular}{c|c||ccc}
    \toprule
    Point cloud config & $E$ layer          & CD $\downarrow$   & ECD $\downarrow$  & IR $\downarrow$ \\ \midrule
    $N\in \mathbb{R}^{4096 \times 3}$  & 6  & 0.17          & 3.80         & 0.97 \\
    $N\in \mathbb{R}^{8192 \times 3}$  & 6  & 0.19          & 3.87         & 0.41 \\
    $N\in \mathbb{R}^{12288 \times 3}$ & 6  & 0.17          & 3.89         & 0.28 \\
    $N\in \mathbb{R}^{8192 \times 3}$  & 12 & 0.16         &  3.53         & 0.28 \\
    \bottomrule
    \end{tabular}
    \label{tab:robust_data}
\end{table}

    \head{Point cloud corruptions and point cloud encoder.}
    We validate the robustness of the method to different point cloud corruptions, like noises, sparsity, and partial point masking, in Fig.~\ref{fig:robust}. We observe that the basic structure of a shape can be reconstructed, however, the local details are not consistent with the input CAD model due to noise or incomplete point clouds.
    We test the robustness of the method to different numbers of points in the input point clouds in Tab.~\ref{tab:robust_data}.
    Results indicate sparse point clouds lead to worse results on geometric metrics. This is because point clouds with fewer points are not able to represent small-scale geometry making the CAD reconstruction with fine-level details difficult. 
    
    We also show the impact of different numbers of Transformer layers in the point cloud encoder. More Transformer layers can increase the representation capacity of the network, thus improving the metrics. 

\begin{figure*}[t]
    \centering
    \includegraphics[width=\textwidth]{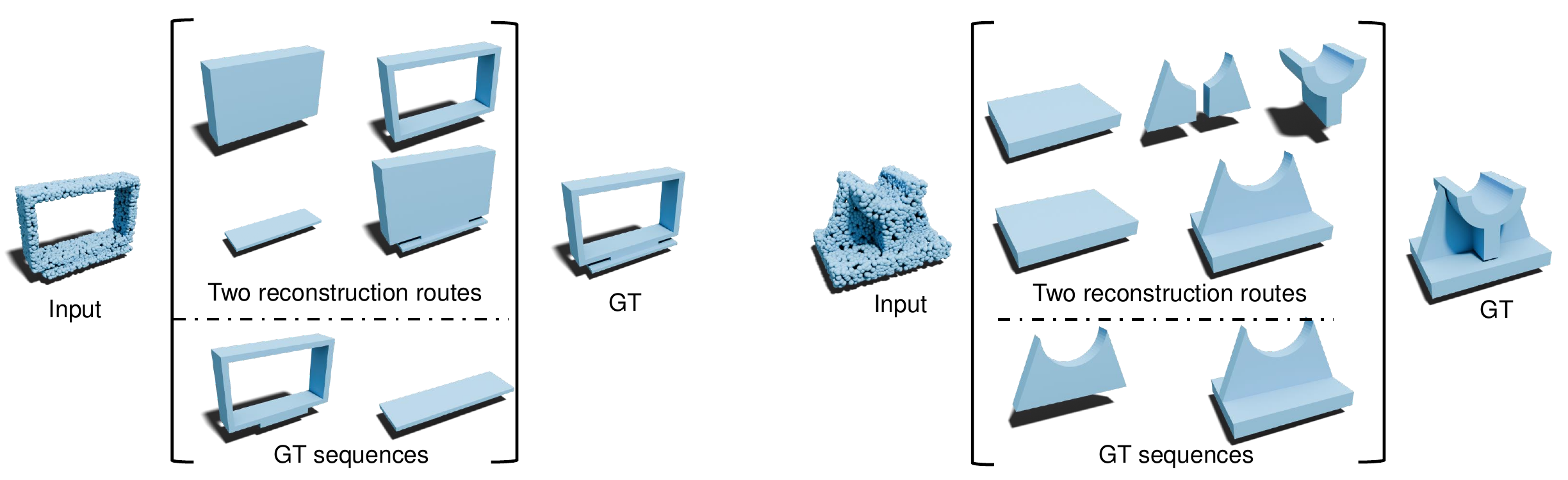}
    \caption{Multiple feasible CAD sequences for a point cloud can be provided to reconstruct an input point cloud.}
    \label{fig:multi-cads}
\end{figure*}

\begin{figure*}[t]
    \centering
    \includegraphics[width=\textwidth]{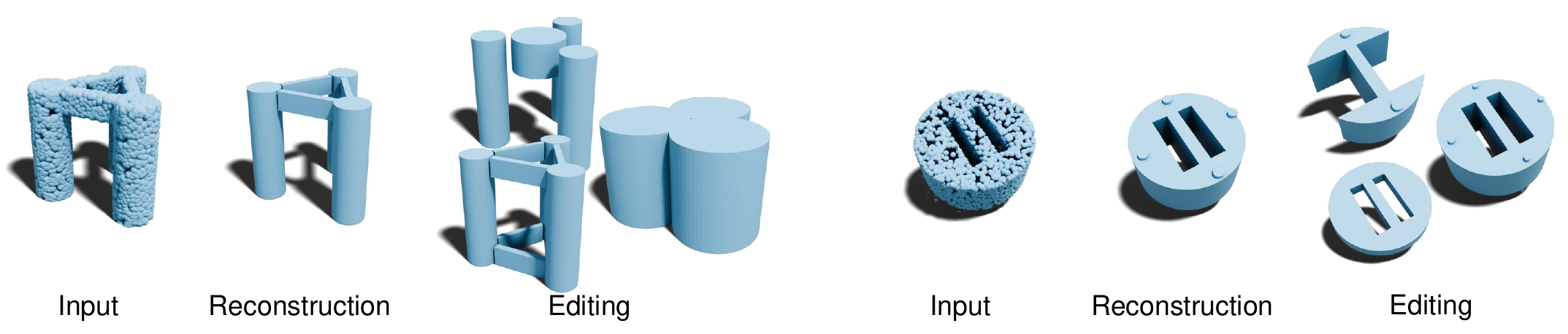}
    \caption{Editing results for reconstructed CAD models.}
    \label{fig:cad-edit}
\end{figure*}

\subsection{Discussions}
    \head{Prompt sampling vs. Token sampling.}
    In Fig.~\ref{fig:ablation-sample-comp}, we show the capability of prompt sampling over the token sampling of the autoregressive model. In HNC-CAD, the sampling results lead to diverse CAD modeling steps, but few of them convey the structure of the input model. In Ours (token), we turn to use token sampling to produce candidate CAD modeling steps and the results are almost the same. With the help or prompt shown in different colors, Ours (step) can produce multiple reasonable candidates. 
    
\begin{figure}
    \centering
    \includegraphics[width=\linewidth]{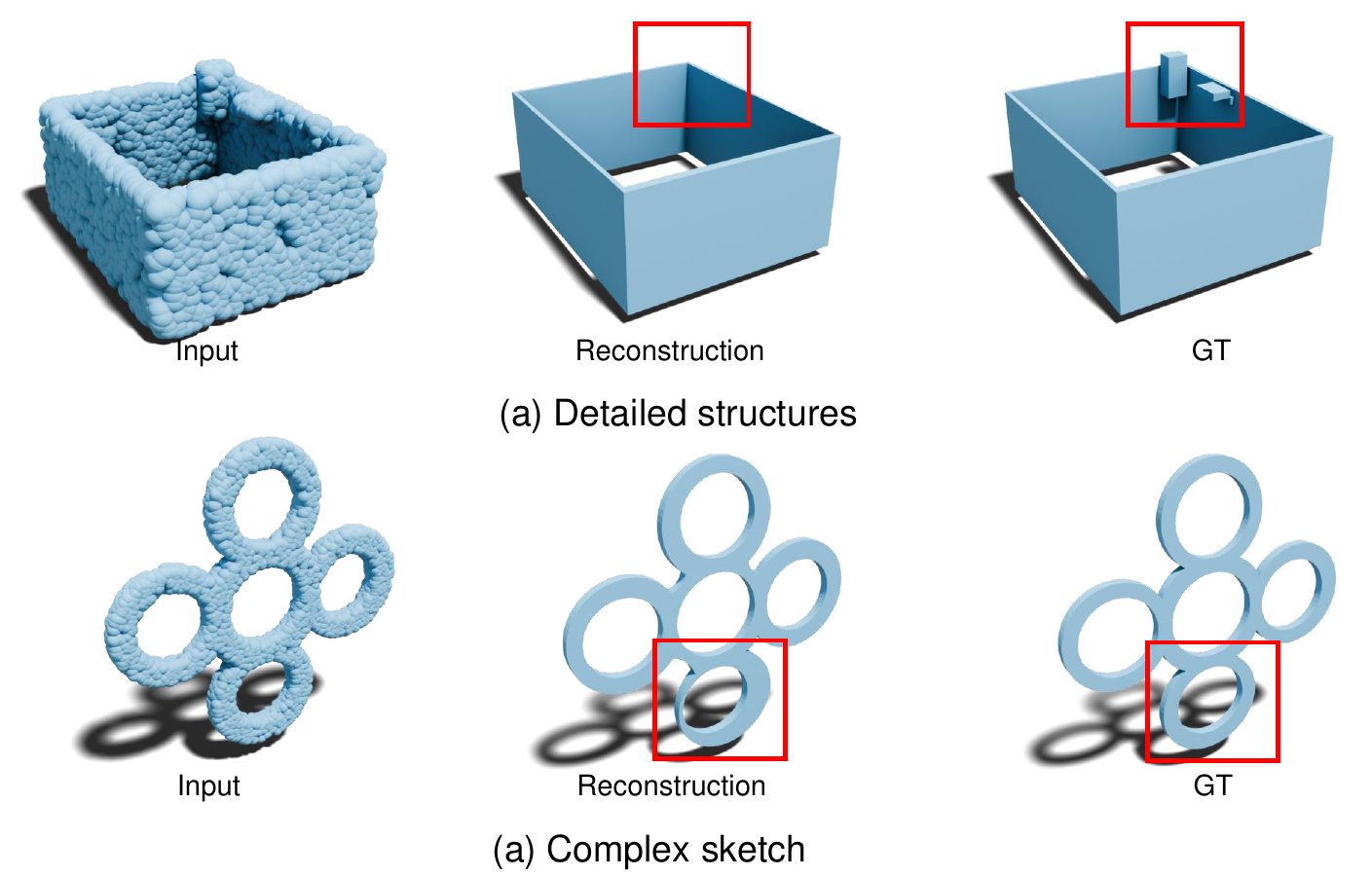}
    \caption{Failure cases. The method has difficulty in recovering (a) details that are very small compared to the overall input shape and (b) complex sketches.}
    \label{fig:fail}
\end{figure}

    \head{Relationship with cross-section-based methods.}
    In geometric guidance computation, we sample planar prompts on RANSAC-detected planes. These planes are potential cross-sections which an extrusion cylinder can start from. As shown in Fig.~\ref{fig:ablation-sample-comp}, by utilizing prompts, our approach can reconstruct the extrusion cylinder highly correlated with the cross-section. From this perspective, our approach is similar to the cross-section-based method Point2Cyl. Different from Point2Cyl which uses a closed-form formulation to calculate extrusion parameters, we use the network to predict the extrusion operation. The network is more robust to the input and produces powerful geometric descriptors for reconstruction.   
    
    \head{Failure cases.}
    In Fig.~\ref{fig:fail}, we demonstrate two typical failure cases. 
    In the first case, the input shape contains an extrusion with a larger size and two small blocks. 
    The method fails to recover the two small blocks as they contain very few points.  
    In the second case, the input shape is extruded from a complex sketch and the method predicts an invalid sketch that is too different from the input.

\subsection{Applications}
    \label{subsec:applications}

    \head{Interactive reconstruction.}
    Our method can produce different reconstruction sequences for the same input. This advantage can be attributed to the promptable reconstruction, which facilitates multiple legitimate candidate reconstructions in each stage. In addition to always selecting the best candidate in the single-step selection, it is also possible to sample according to probabilities. Illustrations are shown in Fig.~\ref{fig:multi-cads}. This sampling can still lead to correct reconstructions.

    Moreover, we can explore potential variations of a reconstructed CAD shape by editing parameters within the reconstructed CAD modeling sequence (e.g., see Figure.~\ref{fig:cad-edit}).

\section{Conclusion and Future Work}
    We present a prompt-and-select framework for sequential CAD modeling from point clouds. 
    The method leverages local geometric information to guide the CAD reconstruction process of an autoregressive model. We provide plane-based prompts for better candidate sketch-extrusion sequence prediction.
    Extensive experiments validate the method is effective and can outperform SoTA methods in geometry accuracy (CD, HD, and ECD metrics), valid ratios, and geometry regularity (NC).
    We also showcase that the method can be applied to reverse engineering and further geometry editing. 
    
    There are some limitations to our work. 
    First, we did not consider the regularity constraints between different sketch-extrusions, e.g., co-axes, alignment, and orthogonal relations between different surfaces.
    Second, the method mainly focuses on the CAD modeling of sketch-extrusion-based shapes and does not deal with operations like sweeping and revolving surfaces. However, it should be possible to extend it to different kinds of CAD commands when corresponding datasets are available for training. 
    Third, fine-grained geometry details usually pose challenges to both our method and previous work. These details are usually under-represented in input point clouds and optimization losses compared to larger geometry primitives. 

\bibliographystyle{ACM-Reference-Format}
\bibliography{references}

\appendix
\section{CAD Domain-specific language Sequence}\label{app:temb}
    The domain-specific language (DSL) defined by SkexGen~\cite{xu2022skexgen} contains the most frequently used CAD modeling operations, including curves \{line, arc, circle\}, feature operation \{extrude\}, and Boolean operation \{union, subtraction\}.
    Tab.~\ref{tab:DSL-spe} provides the overview of the tokens and parameters of the DSL in SkexGen~\cite{xu2022skexgen}.
    SkexGen introduces a set of special tokens to denote different types of CAD modeling operations, where $\varepsilon_*$ marks the end of "[operation type]", e.g. $\varepsilon_c, \varepsilon_e$ for the end of a curve and the end of extrusion.
    The parameters are supposed to be continuous and we follow previous works~\cite{DBLP:conf/iccv/WuXZ21, xu2022skexgen, DBLP:conf/icml/XuJLWF23} to quantize them into discrete values to simplify the auto-regressive learning of the CAD DSL sequence.
    We have Eq.~\ref{eq:dsl-quan} for the quantization,
    \begin{equation}
    \begin{split}
        \mathbf{V} &= \Big \lfloor \frac{v - X_\text{min}}{X_\text{max} - X_\text{min}} \times \mathbf{Q} \Big \rfloor,
    \end{split}
    \label{eq:dsl-quan}
    \end{equation}
    where $v$ is the continuous value of a parameter, $\mathbf{V}$ is its quantization counterpart. $X_\text{min}$ and $X_\text{max}$ are the range of $v$, $\lfloor \cdot \rfloor$ is Floor operation, $\mathbf{Q}$ is the number of quantization bins of the $[X_\text{min}, X_\text{max}]$, i.e. to determine the number of divisions from $X_\text{min}$ to $X_\text{max}$.
    %
    %
    
    %
    As a counterpart, we formulate the value dequantization as Eq.~\ref{eq:dsl-dequan} to convert $\mathbf{V}$ back to a continuous value,
    \begin{equation}
    \begin{split}
        \hat{v} = \mathbf{V} &\times \frac{X_\text{max} - X_\text{min}}{\mathbf{Q}},
    \end{split}
    \label{eq:dsl-dequan}
    \end{equation}
    where the $\hat{v}$ is the dequantized approximation of $v$. The difference between $\hat{v}$ and $v$ is less than $\frac{X_\text{max} - X_\text{min}}{\mathbf{Q}}$.
    
    \begin{table}[h]
        \centering
        \setlength{\tabcolsep}{1mm}
        \caption{Descriptiong of CAD DSL tokens.}
        \resizebox{\linewidth}{!}{
        \begin{tabular}{cccc}
        \toprule
        \makecell{Operation} & \makecell{Token \\ Type} & \makecell{Token \\ Value Range} & \makecell{Description} \\ \midrule
        & $\varepsilon_s$ & 1& End of a Sketch\\ \cmidrule{2-4}
        & $\varepsilon_f$ & 2& End of a Face \\ \cmidrule{2-4}
        Sketch& $\varepsilon_l$ & 3& End of a Loop \\ \cmidrule{2-4}
        & $\varepsilon_c$ & 4& End of a Curve \\ \cmidrule{2-4}
        & $\varepsilon_x$  & $[5,68]$ & Curve   \\ \cmidrule{2-3}
        & $\varepsilon_y$  & $[5,68]$ & Coordinates \\ \midrule
        &$d^+$& $[2, 65]$& \makecell{Extruded Distance Towards\\ Sketch Plane Normal}\\ \cmidrule{2-4}
        &$d^-$& $[2, 65]$& \makecell{Extruded Distance Opposite\\ Sketch Plane Normal}\\ \cmidrule{2-4}
        &$\tau_x$& $[2, 65]$& \\ \cmidrule{2-3}
        Extrude&$\tau_y$& $[2, 65]$& Sketch Plane Origin\\ \cmidrule{2-3}
        &$\tau_z$& $[2, 65]$ & \\ \cmidrule{2-4}
        &$\theta$& $[2, 65]$ & \\ \cmidrule{2-3}
        &$\phi$& $[2, 65]$ & Sketch Plane Orientation\\ \cmidrule{2-3}
        &$\rho$& $[2, 65]$& \\ \cmidrule{2-4}
        &$\sigma$& $[2, 65]$& Sketch Scaling Factor\\ \cmidrule{2-4}
        &$\varepsilon_e$ & 1& End of an Extruding Parameters \\ \midrule
        \thead{Boolean \\  Operations} & $\beta$ & $\{1,0\}$& \{Union, Subtraction\}  \\ \bottomrule
        \end{tabular}
        }
        \label{tab:DSL-spe}
    \end{table}

    The sketch part $s$ is organized as a hierarchy of multiple curves that make up the normalized sketch. Specifically, $s$ represents a "sketch-face-loop" hierarchy, in which a sketch can contain one or more faces, with a face being a 2D region bounded by loops. The loop is a closed path consisting of one or more curves. 
    We use $\varepsilon_s, \varepsilon_f, \varepsilon_l$ to represent the separation of hierarchical elements in $s_\text{cur}$. Different types of curves are uniformly noted as $\varepsilon_c$, identified by different numbers of curve parameters:
    \begin{itemize}
    \item Line--2 operation parameters, Start and End point.
    \item Arc--3 operation parameters, Start, Mid, and End point.
    \item Circle--4 operation parameters, 4 points that connect into mutual orthogonal diameters.
    \end{itemize}
    The curve parameters are 2D coordinates on the XY-axis and will be projected onto the sketch plane through the linear transformation derived from the extrude operation.
    In particular, given an arbitrary point on a curve $\vec{\varepsilon} = (\varepsilon_x, \varepsilon_y, 0)^\mathbf{T}$, the Euler angle $(\theta,\phi,\rho)$, the translation vector $\vec{\tau} = (\tau_x,\tau_y,\tau_z)^\mathbf{T}$, and the scaling factor $\sigma$, the projected point $\vec{\varepsilon_p}$ is given by Eq.~\ref{eq:transformation},
    \begin{equation}
        \vec{\varepsilon_p} = \mathbf{R}_{xyz}(\theta,\phi,\rho)(\sigma \cdot \vec{\varepsilon}) + \vec{\tau}
    \label{eq:transformation}
    \end{equation}
    where $\mathbf{R}_{xyz}(\theta,\phi,\rho) \in \mathcal{SO}(3)$ combines the Euler angles in a rotation matrix in the orthogonal group $\mathcal{SO}(3)$.

\section{Point Cloud Encoder}\label{subapp:pc-encoder}
    We implement the Point-MAE~\cite{DBLP:conf/cvpr/HeCXLDG22} encoder to extract point cloud features. 
    Specifically, the input point cloud is first divided into irregular point patches $p_\text{fps}$ via Farthest Point Sampling (FPS) and K-Nearest Neighborhood (KNN) retrieval. Formally, we have:
    \begin{equation}
    \begin{split}
         p_\text{fps} &= \text{FPS}(p_\text{in}), \quad p_\text{in} \in \mathbb{R}^{N_p \times 3}, \quad p_\text{fps} \in \mathbb{R}^{N_f \times 3}, \\
         p_\text{knn} &= \text{KNN}(p_\text{fps}, p_\text{in}), \quad p_\text{knn} \in \mathbb{R}^{N_f \times k \times 3}, \\
    \end{split}
    \label{eq:app-ge-1}
    \end{equation}
    where $p_\text{in}$ is the input point cloud, FPS samples points $p_\text{fps}$ as centers of $N_f$ point patches, and KNN selects $k$ nearest points from $p_\text{in}$ for each point patch. The embedding of the point patch is calculated as:
    \begin{equation}
    \begin{split}
        p_\text{knn} &= p_\text{knn} - p_\text{fps} \\
        f_\text{knn} &= \text{PointNet}(p_\text{knn}) + \text{PE}(p_\text{fps}), \quad f_\text{knn} \in \mathbb{R}^{N_f \times C},
    \end{split}
    \label{eq:app-ge-2}
    \end{equation}
    where PointNet is a lightweight MLP+max-pooling layer, and PE is a learnable positional embedding also implemented as an MLP layer. $C$ is the number of feature channels.
    $f_\text{knn}$ is then fed into a standard transformer. Each transformer block is implemented as a multi-head self-attention (SA) + MLP layer. The positional embedding $\text{PE}(p_\text{fps})$ is added to each block to provide point position information.
    The output of the transformer is the features of point cloud $p_\text{in}$, denoted as $f_\text{in} \in \mathbb{R}^{N_f \times C}$. To balance cost and performance, we set $k=32$ in KNN, $N_p = 8192, N_f=384, C=384$. The number of transformer blocks is set as 6.

\section{Point Cloud Segmentation Network}
    We train a segmentation network to compare the differing regions between $p_\text{full}$ and $p_\text{prev}$.
    The point cloud segmentation network consists of the Point-MAE encoder, a decoder, and a mask prediction head. 
    The Decoder adopts three Transformer blocks and takes the concatenation of point cloud features $f_\text{full}, f_\text{prev}$ as input and outputs the features $f_\text{mask} \in \mathbb{R}^{2M \times C}$ which is fed to the mask prediction head.
    The mask prediction head is implemented as a fully connected layer to project $f_\text{mask}$ to a point cloud mask $\mathcal{M}$. Then followed by a reshape and sigmoid, the segmentation mask $\mathcal{M}$ is obtained as follows: 
    \begin{equation}
         \mathcal{M} = \ \text{Sigmoid}  (\text{Reshape}( \text{FC}(f_\text{mask})), \ \mathcal{M} \in \mathbb{R}^{2(N_f\times k) \times 1}.
    \label{eq:app-seg}
    \end{equation}
    The first $N_p$ masks in $\mathcal{M}$ are for the differing regions of $p_\text{full}$ w.r.t. $p_\text{prev}$ while the next $N_p$ masks are for the differing region of $p_\text{prev}$ w.r.t. $p_\text{full}$.
    We employ the binary cross-entropy loss to optimize the point cloud segmentation network.

    \head{Discussion.} There are various applicable techniques to obtain the differing regions between $p_\text{full}$ and $p_\text{prev}$.
    For example, a potential approach is first to perform point cloud registration on $p_\text{full}$ and $p_\text{prev}$ to align two point clouds. Followed by calculating the minimum Euclidean distance of each point in one point cloud to the other, we retain those points with distances exceeding a certain threshold.
    We believe that employing an advanced point cloud registration framework (e.g.~\cite{wang2023roreg}) allows us to obtain more precise inputs for our sequence reconstruction network, which we leave for future work.

\section{Single-step Reconstruction and Selection} \label{subapp:seq-dec}
\subsection{Differentiable 3D Bounding Box Contruction} \label{subapp:ssr-geoloss}
    In this section, we first introduce how to construct a 3D bounding box using parameters in a single-step sketch-extrude sequence $o_t$. Then we introduce how to construct the 3D bounding box in a differentiable way using the output of the sketch decoder ($s$-dec) and extrude decoder ($e$-dec).

    Executing a single-step sketch-extrude sequence constructs an extrusion cylinder by extruding a sketch on the sketch plane. By replacing the sketch with its bounding box, the execution can construct the 3D bounding box of the extrusion cylinder while ensuring a differentiable process w.r.t. extruding parameters and sketch parameters.
    We first calculate the 2D bounding box of a sketch by calculating the maximal and minimal values along the x/y axis.
    Given a point representing one corner of the sketch bounding box $\vec{\varepsilon}^{\ l}=(\varepsilon_x^l, \varepsilon_y^l, 0)^\mathbf{T}$ and the extruded distance $d^+, d^-$, we apply Eq.~\ref{eq:transformation} with a slight modification to project $\vec{\varepsilon}^{\ l}$ to two corresponding corners of the 3D bounding box as follows:
    \begin{equation}
    \begin{split}
        \vec{\varepsilon_p}^{l1} &= \mathbf{R}_{xyz}(\theta,\phi,\rho)(\sigma \cdot \vec{\varepsilon}^{\ l} + (0,0,d^+)^\mathbf{T}) + \vec{\tau}, \\
        \vec{\varepsilon_p}^{l2} &= \mathbf{R}_{xyz}(\theta,\phi,\rho)(\sigma \cdot \vec{\varepsilon}^{\ l} + (0,0,d^-)^\mathbf{T}) + \vec{\tau}.
    \end{split}
    \label{eq:app-bbox-transformation}
    \end{equation}
    
    Referring to Eq.~\ref{eq:dec-calculateion}, the output of $s$-dec and $e$-dec is the softmax token probability, requiring a non-differentiable $\arg\max$ operator to convert them into token values. 
    We instead calculate a differential sketch/extrude parameter by accumulating the token probability along all possible token values with their dot products. The calculation for the parameter at index $h$ of the $s$-dec output is formatted as follows:
    \begin{equation}
        s^h = \sum\limits_{j=0}^{\mathbf{Q}_{s}-1} s^{h} \times j,
    \label{eq:app-diff-token-v}
    \end{equation}
    where $[0, Q_s-1]$ is the value range of token $s^h$, and $j$ is the component index of the output probability $s\text{-dec}(\cdot, \cdot)$, within a range of $[0, \mathbf{Q}_s-1]$. Once all tokens of $s$ are calculated, the coordinates of sketch parameters can be obtained through dequantization Eq.~\ref{eq:dsl-dequan}. We select the minimum and maximum coordinates to form the corner of the sketch bounding box.
    Without loss of generality, extruding parameters of an output of the $e$-dec can be obtained in the same way.

    Finally, by combining the above designs in Eq.~\ref{eq:app-diff-token-v},~\ref{eq:dsl-dequan},~\ref{eq:app-bbox-transformation}, we can construct the 3D bounding box of the executed shape of a single-step sketch-extrude sequence in a differentiable way, leading to the $\text{bbox}(\cdot)$ operator in Eq.~\ref{eq:geo-loss}.
    
\begin{figure}
    \centering
    \includegraphics[width=\linewidth]{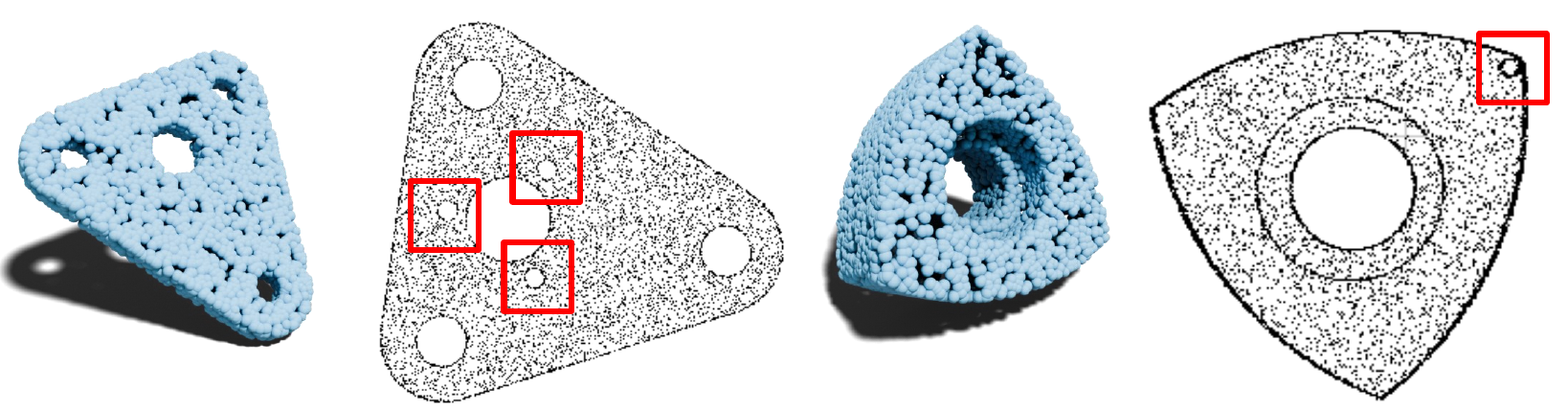}
    \caption{The detailed structure of two point clouds from Fig.~\ref{fig:comparison-Fusion}. There are some holes framed by a red box. Best viewed by zooming in.}
    \label{fig:bird-eye view pcd}
\end{figure}

\subsection{Training implementation} \label{subapp:train}
    \head{Training sample.}
    Our raw training data comes from DeepCAD~\cite{DBLP:conf/iccv/WuXZ21}, where each data item contains a CAD model and one corresponding CAD modeling sequence. 
    The CAD model is obtained by progressively executing the CAD modeling sequence just like Fig.~\ref{fig:teaser}. We sample $p_\text{full}$ from the CAD model and $p_\text{prev}$ can be sampled from the intermediate executions.
    Supposing that a CAD modeling sequence has $N_O$ CAD modeling steps, $p_\text{prev}$ for the first step is an empty point cloud and is identical to $p_\text{full}$ at the final step.

\begin{figure}
        \centering
        \includegraphics[width=\linewidth]{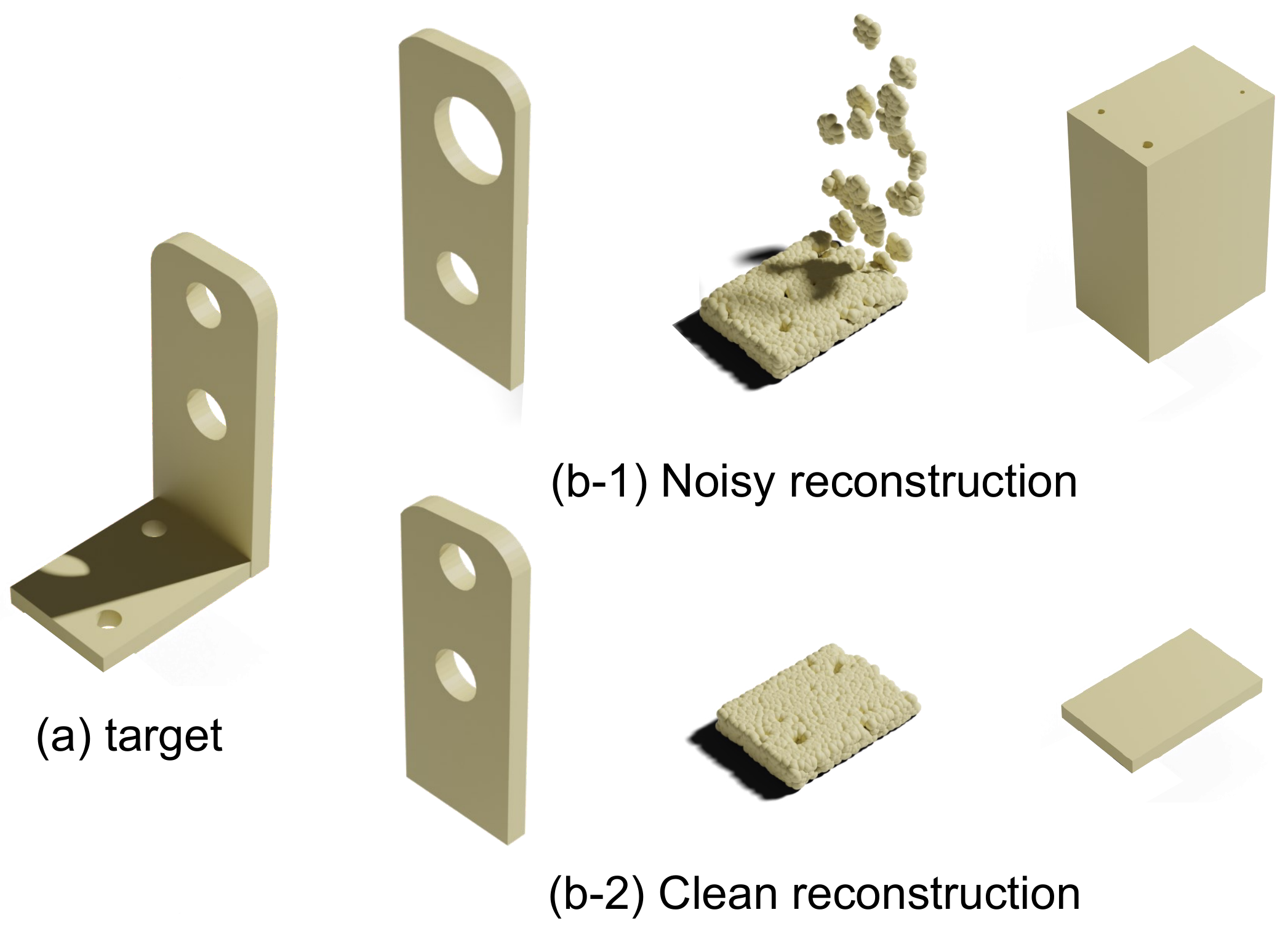}
        \caption{Influence of accumulated noises during CAD modeling. (a) The ground truth 3D shape; (b) reconstruction of previous steps, remaining inputs not yet reconstructed, and reconstruction of the current step. We show results of inaccurate reconstruction of previous steps (b-1) and that of accurate ones (b-2).}
        \label{fig:noise}
    \end{figure}

    \head{Data augmentation.}
    We also notice there are inevitable noises gradually accumulated during the iterative reconstruction, mainly introduced by imperfect point cloud segmentation mask $\mathcal{M}_\text{full}$ and $\mathcal{M}_\text{prev}$ and imprecise sequence reconstruction $o_t$.
    Imperfect $\mathcal{M}_\text{full}$ and $\mathcal{M}_\text{prev}$ leads to outliers or detail losses of a point cloud, while imprecise sequence parameters, e.g., Sketch Plane Origin $(\tau_x, \tau_y, \tau_z)$ in Extruding operations, will induce uncertainty for the executed shape of successive steps.
    They can deteriorate the performance by interfering with the quality of $f_\text{ref}$ and prompt sampling. Even worse, noise accumulation may lead to reconstruction failure, as depicted in Fig.~\ref{fig:noise}.
    
    To mitigate the adverse effects of noises, we employ data augmentation to inject noises into training samples, making the network training more robust.
    Specifically, the noise of training samples is induced in the $\mathcal{M}_\text{full}$ and $\mathcal{M}_\text{prev}$ and training prompts. 
    Noises introduced into $p_\text{ref}$ can be controlled by $\mathcal{M}$, and we randomly invert $\mathcal{M}_\text{full}$ or $\mathcal{M}_\text{prev}$ with a probability of $5 \%$.
    Second, training prompts sampled from sketch planes are scaled with a factor ranging from $[0.8, 1.1]$. In this way, prompts can adapt to noisy inputs not perfectly aligned with the ground-truth sketch plane.

\end{document}